%% file: 0_main.tex
\documentclass[11pt]{article}
\pdfoutput=1

\usepackage[final]{acl}

\usepackage{times}
\usepackage{latexsym}

\usepackage[T1]{fontenc}

\usepackage[utf8]{inputenc}

\usepackage{microtype}

\usepackage{inconsolata}

\usepackage{graphicx}
\usepackage{tabularx}
\usepackage{booktabs}
\usepackage{multirow}
\usepackage[table]{xcolor}
\usepackage{pifont}
\newcommand{\cmark}{\ding{51}}
\newcommand{\xmark}{\ding{55}}
\newcommand{\abilityyes}{\cellcolor{gray!15}\cmark}
\usepackage{algorithm}
\usepackage{algpseudocode} 
\usepackage{multicol}
\usepackage[most]{tcolorbox}
\usepackage{hyperref}
\hypersetup{
  pdftitle={EMemBench: Interactive Benchmarking of Episodic Memory for VLM Agents},
  pdfauthor={Xinze Li, Ziyue Zhu, Siyuan Liu, Yubo Ma, Yuhang Zang, Yixin Cao, Aixin Sun},
}
\usepackage[capitalize]{cleveref}
\usepackage{xspace}
\usepackage{comment}
\usepackage{placeins}

\newcommand{\benchname}{EMemBench\xspace}

\makeatletter
\newcommand{\StatexIndent}[1][1]{%
  \setlength\@tempdima{\algorithmicindent}%
  \Statex\hskip\dimexpr#1\@tempdima\relax}
\makeatother
\usepackage[dvipsnames]{xcolor}
\newcommand{\CaseTitle}[1]{\vspace{2pt}\noindent\textbf{#1}\par}
\newcommand{\Qline}[1]{\textcolor{NavyBlue}{\textbf{Question:}}~\textcolor{NavyBlue}{#1}}
\newcommand{\GTline}[1]{\textcolor{ForestGreen}{\textbf{GT:}}~\textcolor{ForestGreen}{\texttt{#1}}}
\newcommand{\Predline}[1]{\textcolor{BrickRed}{\textbf{Pred:}}~\textcolor{BrickRed}{\texttt{#1}}}
\newcommand{\Expline}[1]{\textcolor{black!75}{\textbf{Model explanation:}}~\textcolor{black!75}{#1}}
\newcommand{\Ctxline}[1]{\textcolor{TealBlue}{\textbf{Relevant context:}}~{#1}}
\newcommand{\Ctxblock}[1]{%
  \textcolor{TealBlue}{\textbf{Relevant context:}}\\[-2pt]
  \noindent\hspace*{0.8em}%
  \begin{minipage}[t]{0.95\linewidth}
    \footnotesize\ttfamily\raggedright
    #1
  \end{minipage}\par
}

\title{EMemBench: Interactive Benchmarking of \\Episodic Memory for VLM Agents}

\author{
Xinze~Li$^{1,3}$,
Ziyue~Zhu$^{2}$,
Siyuan~Liu$^{2}$,
Yubo~Ma$^{1}$,
Yuhang~Zang$^{3}$,
Yixin~Cao$^{2\dagger}$,
Aixin~Sun$^{1\dagger}$ \\
\\
$^{1}$ Nanyang Technological University, Singapore \quad
$^{2}$ Fudan University, Shanghai, China \\
$^{3}$ Shanghai AI Laboratory, Shanghai, China \\
\\
$^{\dagger}$ Corresponding authors
}

\begin{document}
\maketitle
\begin{abstract}
We introduce \benchname, a programmatic benchmark generator for evaluating long-term episodic memory of agents through interactive games. Rather than using a fixed set of questions, \benchname generates questions from environment-grounded trajectories, covering both text-only and visual game environments. Each template computes verifiable ground truth from underlying game signals, with controlled answerability and balanced coverage over memory skills: single/multi-hop recall, induction, temporal, spatial, logical, and adversarial.  We evaluate memory agents with strong LMs/VLMs as backbones, using in-context prompting as baselines. Across 15 text games and multiple visual seeds, results are far from saturated: induction and spatial reasoning are persistent bottlenecks, especially in visual settings. Persistent memory yields clear gains for open backbones on text games, but improvements are less consistent for VLM agents, suggesting that visually grounded episodic memory remains an open challenge. A human study further contextualizes the difficulty and interpretability of \benchname.

\end{abstract}

\input{1_introduction}
\input{2_relatedwork}
\input{3_benchmark_construction}
\input{4_experiment}
\input{5_analysis}
\input{6_conclusion}

\section*{Limitations}
\paragraph{Limited environment coverage.}
While we evaluate on both text-only (Jericho-style interactive fiction) and visual (Crafter) settings, the number of game environments is still limited. This is largely because our question construction is \emph{environment-dependent}: Single-/Multi-Hop question templates can transfer to different game environments relatively well, but reasoning question templates (e.g., spatial/temporal reasoning) require deeper, game-specific understanding and careful validation, making it costly to scale to other games. 

\paragraph{Contamination is reduced but not eliminated.}
Our questions are generated from an agent’s own interaction trajectory rather than a fixed static dataset, which helps mitigate test-set leakage. However, the generation pipeline itself is algorithmic and fixed, so we cannot fully rule out contamination via pattern/template memorization or distributional overlap as models and training corpora evolve.

\paragraph{Small-scale human study.}
Our human evaluation involves 11 annotators for text games and only 3 for the visual game, so the visual-game human results should be read as indicative rather than definitive. Expanding the annotator pool, especially for visual environments, is left to future work.

\paragraph{Limited model and agent coverage.}
We only evaluate a small set of representative backbones and memory agents. Different memory designs (e.g., indexing, consolidation, retrieval budgets) can change the trade-offs substantially, and broader coverage is needed for more robust conclusions. We plan to release a public leaderboard and continuously expand evaluated models, memory agents and settings.

\section*{Ethical Statement}
While our goal is measurement, stronger long-term memory can enable harmful applications. We position the benchmark as a diagnostic tool and encourage reporting together with safeguards such as data minimization, transparency about what is stored, and deletion controls, rather than optimizing retention indiscriminately. 

In addition, we exclude game environments with explicit or graphic violence; however, some included games still contain highly stylized, game-typical fantasy elements (e.g., enemies such as zombies/skeletons, magic, and combat verbs like \emph{kill}). These depictions are non-realistic and occur in a clearly game-like context (e.g., Crafter’s survival setting requires defending against hostile skeletons; Jericho consists of human-authored interactive fiction including topics like fantasy or dungeon).

\section*{Acknowledgments}
This work was supported by Shanghai Artificial Intelligence Laboratory. This research is supported by cash and in-kind funding from NTU S-Lab and industry partner(s). This work is supported by OPPO.


\bibliography{custom}

\appendix
\input{7_appendix}

\end{document}

%% file: 1_introduction.tex
\section{Introduction}
\label{sec:intro}

\begin{figure}[t]
  \centering
  \includegraphics[width=\linewidth]{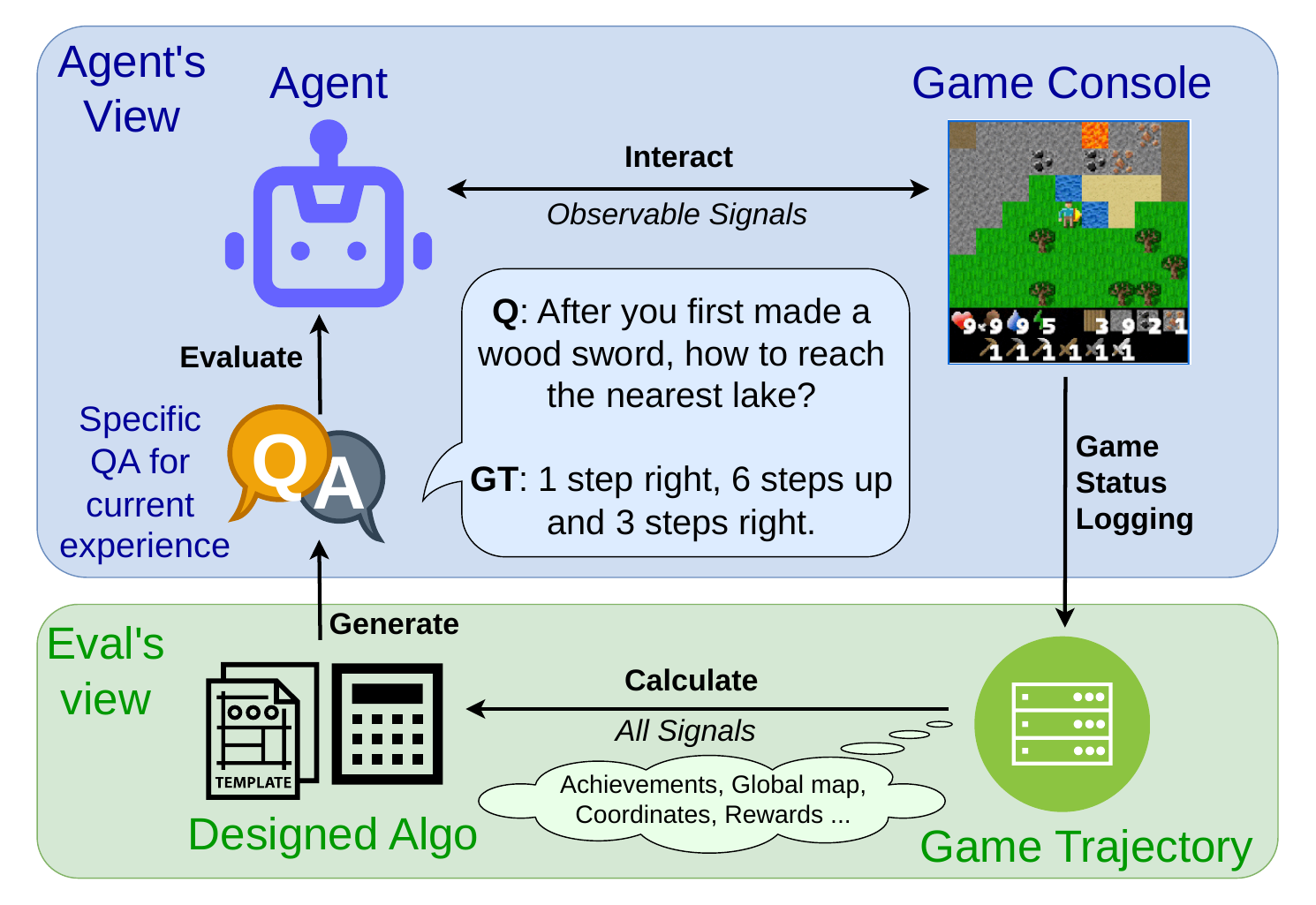}
  \caption{\textbf{\benchname overview.} An agent interacts with a game environment; we log agent-observable and underlying game signals, convert each episode into a QA set with computed ground truths, and the same agent answers using only agent-observable context plus its own memory.}

  \label{fig:intro_fig}
\end{figure}

Large language models (LLMs)~\cite{brown2020language} are increasingly deployed as interactive agents~\cite{survey_agent}, such as web/OS assistants~\cite{osagents} and tool-using copilots~\cite{toolformer}. To make coherent long-term decisions in dynamic environments, an agent must continuously retain, update, and leverage information acquired earlier. Yet evaluation practice remains largely static: many benchmarks assess how well models answer questions given a pre-defined history generated from fixed conversation scripts. Such evaluation provides valuable evidence for long-context reasoning and retrieval, but does not treat memory as an interactive and individualized competence.

Cognitive psychology views human memory as a set of processes for encoding, storing, and retrieving information over time~\cite{melton1963memory, sternberg1999cognitive, matlin2005cognition}, and distinguishes semantic memory---general knowledge---from episodic memory, which records personal experiences and is therefore inherently individualized~\cite{greenberg2010interdependence, tulving1973encoding}. We borrow this distinction purely as a guideline for \emph{what to measure}, without any claim that LLM agents implement human-like memory processes: evaluating episodic memory means testing how well an agent can form and access a record of \textit{its own experience}, not merely how well it can reason over a long document provided by someone else.

Existing benchmarks~\cite{MemoryBank,locomo} abstract away the memory formation process during interaction, and thus do not measure memory as an experience-conditioned capability (Section~\ref{sec:related}).
Motivated by both practical agent deployments and the design guideline above, we argue that an ideal benchmark for agent memory should satisfy three properties.
(i) Individualized: each agent is evaluated on memories from its own experiences and decisions.
(ii) Automatic and scalable: because each agent's interaction history emerges on the fly during evaluation, questions must be generated automatically---manually annotating every new trajectory would be prohibitively expensive.
(iii) Accurate and verifiable: ground-truth answers must be correct and verifiable without human post-editing.

To this end, we introduce \benchname, an experience-conditioned memory evaluation framework (Figure~\ref{fig:intro_fig}) using interactive games as controllable environments.\footnote{We release the generator code, environment configurations, and the frozen controlled-exposure trajectories and QA sets at \url{https://github.com/InternLM/EMemBench}.}
Games not only induce individualized experience for each agent through its observations and decisions, but also expose underlying game signals (e.g., rewards, global maps) for precise and programmatic ground-truth computation.
Building on both text-only and visual environments, each run in the primary setting proceeds in three phases: (i) the agent first plays an episode to produce a trajectory, (ii) a generator then turns this episode trace into a balanced set of questions with deterministic ground truth, and (iii) we run evaluation end-to-end: the same agent that plays the episode answers the questions, using the memory it formed during interaction. In addition, \benchname includes a complementary controlled-exposure setting in which all methods are evaluated on the same fixed interaction history. The two settings are complementary: self-experience evaluation tests whether an agent can form and use memory \emph{while acting}, whereas controlled exposure removes trajectory variance and isolates how well different memory mechanisms reconstruct the \emph{same} history---so systematic gaps between the two are themselves diagnostic (Section~\ref{sec:controlled_exposure_eval}).

However, experience-conditioned evaluation introduces a fairness challenge: when agents follow different trajectories, their question instances cannot be identical. To keep scores comparable, we fix the template library, generation recipe, and random seeds; enforce balanced coverage across ability categories; verify stability across environment seeds; and introduce query-horizon control, which restricts evidence selection and ground-truth computation to a fixed window of interaction steps to remove length-based confounds. Together with structured state logging and controlled answerability preconditions, these controls make individualized, fully automated, and verifiable memory evaluation feasible in dynamic interactive environments.

%% file: 2_relatedwork.tex
\section{Related Work}
\label{sec:related}

\input{8_dataset_table}

Existing benchmarks and systems for long-term memory in LLM agents can be broadly grouped into three lines: (i) static long-history QA benchmarks, (ii) dynamic or interactive memory benchmarks, and (iii) memory architectures that explicitly construct structured world models. Table~\ref{tab:longterm-memory-benchmarks-compact} summarizes representative benchmarks along the dimensions most relevant to our design: modality, scale, construction pipeline, interactivity, and ability coverage.

\paragraph{Static long-history memory benchmarks.}
A large body of prior work evaluates memory by asking questions over a fixed, pre-built history: multi-day personalized conversations in MemoryBank~\cite{MemoryBank}, personalized long-term QA in PerLTQA~\cite{PerLTQA}, very long multi-session dialogues in LoCoMo~\cite{locomo} and BEAM~\cite{beam}, and a curated ability taxonomy over assistant chat in LongMemEval~\cite{longmemeval}. These benchmarks enable standardized, fine-grained ability analysis, but as Table~\ref{tab:longterm-memory-benchmarks-compact} shows, they share two structural properties: every system is evaluated on the same pre-authored history, and question construction relies on human annotation or verification that does not scale to new histories. Both conflict with evaluating memory as an experience-conditioned capability, where the history itself depends on the agent's own actions.

\paragraph{Dynamic and interactive memory benchmarks.}
More recent work moves toward dynamic settings in which memory is tested under sequential interaction. MemoryAgentBench~\cite{MemoryAgentBench} converts long inputs into incremental multi-turn interactions, but the underlying content is still fixed in advance. StoryBench~\cite{storybench} stress-tests long-term memory through sequential decisions in a manually transcribed and annotated branching narrative---one of the closest settings to ours---yet it builds on a single pre-authored game rather than generating QA from arbitrary agent trajectories. MANGO~\cite{mango} evaluates mapping and navigation in text games via fixed walkthroughs and spatial QA. Compared with these, \benchname generates questions from each agent's \emph{own} trajectory, covers seven ability categories beyond spatial reasoning, extends to visual environments, and supports both self-experience and controlled-exposure evaluation under the same generator.

\paragraph{Structured memory and world-model approaches.}
Another related line studies how agents should represent and maintain memory, rather than primarily how to benchmark it: MemBench~\cite{membench} evaluates memory agents along effectiveness, efficiency, and capacity; MemoryBench~\cite{MemoryBench} studies memory under continual learning with simulated user feedback; AriGraph~\cite{arigraph} builds a memory graph as a world model for interactive agents; and Graph Rectification~\cite{graphrectification} incrementally repairs spatial memory graphs from navigation histories. These systems are natural evaluation subjects for \benchname, which measures how well such architectures actually remember their experienced trajectories, independently of any particular memory design.

In summary, prior benchmarks either fix the history, fix the questions, or fix the memory architecture under study. To our knowledge, \benchname is the first benchmark generator that is simultaneously interactive, trajectory-conditioned, and programmatically verifiable: QA instances are computed directly from environment-grounded trajectories and game state, scaling to arbitrary new trajectories without human annotation while keeping ground truth deterministic.

%% file: 8_dataset_table.tex
\begin{table*}[t]
\small
\setlength{\tabcolsep}{3pt}
\renewcommand{\arraystretch}{1.12}
\begin{tabularx}{\textwidth}{l X c r r r c c c c c c c}
\toprule
\multirow{2}{*}{Benchmark} &
\multirow{2}{*}{Environment} &
\multirow{2}{*}{MM.} &
\multicolumn{3}{c}{Scale} &
\multirow{2}{*}{Pip.} &
\multirow{2}{*}{Inter.} &
\multicolumn{5}{c}{Abilities} \\
\cmidrule(lr){4-6}\cmidrule(lr){9-13}
& & & \#Env & \#Traj & \#Q/Traj & & & IE & MH & SR & TR & AD \\
\midrule
MemoryBank~\cite{MemoryBank} &
Personal companion chat &
\xmark &
1 & 15 & 12.9 &
H & \xmark &
\abilityyes & \abilityyes & & & \\
LoCoMo~\cite{locomo} &
Very long multi-modal chat &
\cmark &
1 & 50 & 150.2 &
M & \xmark &
\abilityyes & \abilityyes & & \abilityyes & \abilityyes \\
PerLTQA~\cite{PerLTQA} &
Chinese personal QA &
\xmark &
1 & 3409 & 2.5 &
M & \xmark &
\abilityyes & \abilityyes & & & \\
LongMemEval~\cite{longmemeval} &
Task-oriented assistant chat &
\xmark &
1 & 500 & 1.0 &
H & \xmark &
\abilityyes & \abilityyes & & \abilityyes & \abilityyes \\
BEAM~\cite{beam} &
Multi-domain long chat &
\xmark &
1 & 100 & 20.0 &
M & \xmark &
\abilityyes & \abilityyes & & \abilityyes & \abilityyes \\
MemBench~\cite{membench} &
Assistant-style user agents &
\xmark &
1 & $\sim$65k & 0.8 &
A & \xmark &
\abilityyes & \abilityyes & & \abilityyes & \abilityyes \\
MemoryAgentBench~\cite{MemoryAgentBench} &
Multi-task text environments &
\xmark &
14 & 130 & 15.9 &
M & \xmark &
\abilityyes & \abilityyes & & & \abilityyes \\
MemoryBench~\cite{MemoryBench} &
User-feedback continual-learning tasks &
\xmark &
11 & $\sim$20k & 1.0 &
A & \xmark &
\abilityyes & \abilityyes & & \abilityyes & \abilityyes \\
StoryBench~\cite{storybench} &
Interactive fiction game &
\xmark &
1 & 80+ & -- &
H & \cmark &
\abilityyes & \abilityyes & & \abilityyes & \abilityyes \\
\textbf{\benchname} &
Text + visual games &
\cmark &
16 & $\infty$ & 80 &
A & \cmark &
\abilityyes & \abilityyes & \abilityyes & \abilityyes & \abilityyes \\
\bottomrule
\end{tabularx}
\caption{
A comparison of long-term memory benchmarks.
\textbf{MM.}: multi-modal; \textbf{Pip.}: QA-construction pipeline (H = human annotated, A = automatic, M = mixed);
\textbf{Inter.}: interactive and individualized evaluation.
\textbf{Scale}: \#Env = number of data sources/environments; \#Traj = number of distinct contexts/trajectories; \#Q/Traj = average questions per trajectory (``--'' if not QA-based).
\textbf{Abilities}: IE = information extraction (single-hop); MH = multi-hop; SR = spatial; TR = temporal; AD = adversarial/conflict.
\benchname additionally covers induction and logical abilities, which are omitted from the table because no compared benchmark targets them.}
\label{tab:longterm-memory-benchmarks-compact}
\end{table*}

%% file: 3_benchmark_construction.tex
\section{Benchmark Construction}
\label{sec:benchmark_construction}

\benchname is not a fixed set of questions, but a benchmark generator: a programmatic pipeline that produces evaluation instances from interaction histories grounded in interactive environments. Concretely, we build on two environments: Jericho for human-authored interactive fiction (text-only) and Crafter for an open-world survival game with visual observations. Given an interaction trace (agent--environment trajectory) and underlying game state, our generator converts them into a suite of memory questions with verifiable ground truth. This design supports two complementary evaluation modes: \textit{self-experience evaluation}, where questions are generated from the agent's own trajectory, and \textit{controlled-exposure evaluation}, where all methods are evaluated on the fixed interaction history.

\subsection{Program-as-a-Benchmark: Task Definition and Interfaces}
\label{subsec:benchmark_program}

An \emph{interaction episode} is a sequence of traces $\tau = \{(o_t, a_t, r_t)\}_{t=1}^{T}$ produced by an agent interacting with an environment over $T$ timesteps, where $o_t$ is the observation (e.g., text in Jericho; image in Crafter), $a_t$ is the agent's chosen action, and $r_t$ is the environment reward. We also record underlying game status $\mathcal{S}$ such as rewards, player coordinates, and global map.

A \emph{benchmark generator} $\mathcal{G}: (\tau, \mathcal{S}) \mapsto \mathcal{Q} = \{q_i, y_i, m_i\}_{i=1}^{N}$ maps an episode (with game state) to a set of QA instances, where $q_i$ is a natural-language question, $y_i$ is the ground-truth answer, and $m_i$ is question metadata (type, evidence pointers, and difficulty). The same generator can be applied either to each tested agent's own episode---making the benchmark experience-conditioned---or to a shared fixed episode, enabling both individualized and controlled-exposure evaluation under a unified framework. Accordingly, the core artifact we release is not ``a dataset'' but (i) the generator code $\mathcal{G}$, (ii) environment configurations (games, hyper-parameters, and seeds), and (iii) an evaluation framework that executes $\mathcal{G}$ consistently and reports metrics.

\subsection{Game State Collection}
\label{subsec:trace_and_state}

\paragraph{Structured logging.}
For every timestep, we store a JSON record with the timestep index, raw observation $o_t$, selected action $a_t$, and episode-level identifiers (game name, seed, run id). Per environment, we additionally log: \textit{text games (Jericho)}---score/moves, location strings, inventory lists, and candidate actions; \textit{visual game (Crafter)}---health stats, inventory, nearby entities, player position, player view, dynamic map, achievements, and events.

\paragraph{State reconstruction.}
From the raw game trajectory, we build two derived views for question and ground truth generation:

\textit{Timeline index} $\mathcal{I}$: fast access to ``what happened at step $t$'' (action, observation, event, agent reason,  and any logged structured fields).

\textit{Event index} $\mathcal{E}$: higher-level events extracted from $\mathcal{I}$ (inventory changes, first/second/last occurrence of events, repeated interactions, location transitions, achievement unlocks and so on).

These indices allow templates to query game history with efficiency and accuracy. $\mathcal{E}$ lets us anchor questions to semantically meaningful events.

\subsection{Programmatic QA Generation}
\label{subsec:qa_generation}

\begin{table}[t]
  \centering
  \scriptsize
  \setlength{\tabcolsep}{3pt}
  \renewcommand{\arraystretch}{1.08}
  \begin{tabular}{@{}l p{0.72\columnwidth}@{}}
    \toprule
    \textbf{Ability} & \textbf{Example question} \\
    \midrule
    Single-Hop  & How many wood\_sword did you have at step 120? \\
    Multi-Hop   & What was the action 4 steps before the first step whose action is SLEEP? \\
    Induction    & What was the longest consecutive run of MOVE\_UP? \\
    Spatial     & At step 103, in which direction was the nearest lake? \\
    Temporal    & Did your food value first fall below 7 happen before you first collect wood? \\
    Logical     & What was the cause of your death in this episode? \\
    Adversarial & At which step did you first collect diamond? \\
    \bottomrule
  \end{tabular}
  \caption{Ability types and example questions.}
  \label{tab:ability_examples}
\end{table}

\paragraph{Question templates.}
We implement a library of templates grouped by reasoning requirements. Each template is a small program that:
(1) selects one or more evidence timesteps from $\mathcal{I}$ and/or $\mathcal{E}$,
(2) constructs a question $q$ in natural language,
(3) computes a deterministic ground-truth answer $y$ from the underlying game signals,
and (4) generates metadata $m$ describing the question.

We organize templates into seven categories to test agents’ memory reasoning abilities. We list each ability with one example in Table~\ref{tab:ability_examples}. The full question template sets are provided in Appendix~\ref{app:questiontemplate}.

\paragraph{Answerability controls.}
Each template defines explicit preconditions for being answerable. For answerable questions, we instantiate template variables using values supported by the trajectory and game state (e.g., for ``Have you made a [value]?'', the inserted value is chosen from items valid under the current episode history). We also construct adversarial questions by deliberately violating these preconditions, yielding controlled unanswerable or counterfactual instances---answerability is thus a programmatically controlled property rather than an accidental byproduct of sampling.

\paragraph{Question distribution balance.}
To prevent the benchmark from being dominated by a few types of questions, we explicitly balance the distribution of QA instances across ability categories. A generation parameter (\texttt{max\_per\_type}) caps the number of instances per template per trajectory. With our default \texttt{max\_per\_type}=2, the template inventory (39 non-adversarial templates for text games, 37 for the visual game; Appendix~\ref{app:questiontemplate}) plus adversarial instantiations yields roughly 80 questions per trajectory, with per-category counts capped at twice the template count subject to answerability preconditions.

\paragraph{Language variation.}
To reduce lexical memorization and increase robustness, we optionally apply paraphrasing to a subset of generated questions. Since paraphrasing is applied after ground-truth computation, it cannot change the answer as long as the paraphrase preserves semantics. To filter paraphrase failures, we run automatic consistency checks: paraphrases must preserve all numbers and step indices verbatim, retain domain terms (actions, resources, terrains), and remain a single English question; violations are discarded and the original phrasing is kept.

\paragraph{Adversarial negatives.}
For adversarial questions, we generate hard distractors by sampling plausible alternatives (e.g., confusable items, unreached locations, or not-happened events) from environment-specific vocabularies and from entities appearing in the same episode. This yields counterfactual questions that are syntactically plausible but confounding, probing whether a system overgeneralizes from priors instead of retrieving the specific episodic fact.



\subsection{Fairness and Reproducibility Controls}
\label{subsec:controls}

\paragraph{Controlled-exposure evaluation.}
To complement self-experience evaluation where each method interacts with the environment, \benchname also supports a controlled-exposure setting in which all methods are evaluated on the same fixed interaction history and the same trajectory-derived QA set. In this mode, the input remains an environment-grounded interaction trace, but exposure differences across methods are reduced. This provides a more direct comparison of how different memory mechanisms reconstruct and use the same experienced history.

\paragraph{Query-horizon control.}
Different agents can produce trajectories of varying lengths due to exploration strategies or early termination, which can complicate comparison if question generation always targets the full episode. To reduce this confound, we introduce a query horizon setting that restricts both \textbf{evidence selection} and \textbf{answer computation} to a prefix window of the interaction, from step 1 to a user-specified $N$. When enabled, templates generate questions whose referents are explicitly scoped to this window (e.g., ``From step 1 to 50, how many times did you enter the kitchen?''). Ground-truth answers are recomputed from the same truncated window. We report results under query-horizon control in \cref{queryhorizon}.

\paragraph{Evaluation metrics.}
\textbf{Accuracy} is the mean per-question score under type-specific matching rules: exact match for integers, adaptive-precision or relative-tolerance match for floats, ANLS-based soft match for strings, and maximum over candidates for list-valued references (so questions with multiple acceptable answers are credited if any is produced). \textbf{F1} measures answerability calibration by treating \texttt{not answerable} as the negative class: precision is computed over questions the system attempts to answer, recall over questions that are actually answerable, and F1 is their harmonic mean. Full scoring rules are given in Appendix~\ref{app:score_rules}.

\paragraph{Reproducibility.}
We ensure reproducibility by fixing the environment random seed in experiments. In text games, the seed affects triggered random events; in the visual game, it determines the generated map and the spawning of entities such as skeletons. In our experiments we use five fixed seeds, \{1, 42, 43, 100, 123\}, for visual games and report results averaged over them. For text games, we use seed 42 and average over 15 games. For question generation, we use seed 42. Results are stable across seeds as shown in \cref{randomseed}.

%% file: 4_experiment.tex
\begin{table*}[t]
    \centering
    \scriptsize
    \setlength{\tabcolsep}{2pt}
    \resizebox{\textwidth}{!}{%
    \begin{tabular}{ll*{9}{c}}
        \toprule
        \multirow{2}{*}{Model} & \multirow{2}{*}{Method} &
        \multicolumn{1}{c}{Single-Hop} &
        \multicolumn{1}{c}{Multi-Hop} &
        \multicolumn{1}{c}{Induction} &
        \multicolumn{1}{c}{Spatial} &
        \multicolumn{1}{c}{Temporal} &
        \multicolumn{1}{c}{Logical} &
        \multicolumn{1}{c}{Adversarial} &
        \multicolumn{2}{c}{Overall} \\
        \cmidrule(lr){3-3}
        \cmidrule(lr){4-4}
        \cmidrule(lr){5-5}
        \cmidrule(lr){6-6}
        \cmidrule(lr){7-7}
        \cmidrule(lr){8-8}
        \cmidrule(lr){9-9}
        \cmidrule(lr){10-11}
        & &
        ACC &
        ACC &
        ACC &
        ACC &
        ACC &
        ACC &
        ACC &
        ACC & F1 \\
        \midrule
        \multicolumn{11}{l}{\textbf{Text-only Games}} \\
        \midrule
        \multirow{4}{*}{GPT-5.1}
            & In-context   & \textbf{76.7} & \textbf{35.7} & 30.7 & 34.5 & 54.1 & 57.6 & 46.9 & 49.7 & \textbf{48.3} \\
            & Mem0         & 43.4 & 8.7 & 23.0 & 13.5 & 53.5 & 18.4 & \textbf{92.1} & 36.3 & 33.7 \\
            & LangMem      & 42.7 & 33.0 & \textbf{34.2} & \underline{48.1} & \textbf{70.1} & 23.5 & 87.9 & 48.8 & 44.7 \\
            & A-MEM        & 51.4 & \underline{34.6} & 21.0 & 46.0 & 62.7 & \underline{63.2} & 73.5 & 49.9 & 46.8 \\
        \cmidrule{1-11}
        \multirow{4}{*}{Qwen2.5-32B-Instruct}
            & In-context   & 52.3 & 26.0 & 23.1 & 35.3 & 53.2 & 50.0 & 49.1 & 40.8 & 39.9 \\
            & Mem0         & 51.4 & 5.8 & 11.6 & 12.3 & 40.7 & 23.9 & \underline{90.4} & 35.2 & 30.7 \\
            & LangMem      & 49.0 & 29.7 & \underline{33.8} & \textbf{50.4} & 55.5 & 30.9 & 85.3 & 49.0 & 43.7 \\
            & A-MEM        & 66.0 & 27.2 & 17.2 & 37.4 & 58.2 & \textbf{65.8} & 88.7 & \underline{51.4} & \underline{48.1} \\
        \cmidrule{1-11}
        \multirow{4}{*}{Qwen3-32B}
            & In-context   & 49.0 & 32.7 & 25.6 & 46.2 & \underline{64.7} & 38.2 & 54.4 & 44.9 & 41.8 \\
            & Mem0         & 50.6 & 5.0 & 33.3 & 8.9 & 50.8 & 27.1 & 84.7 & 41.4 & 37.0 \\
            & LangMem      & 47.3 & 17.9 & 21.5 & 45.1 & 57.6 & 29.9 & 71.1 & 42.2 & 36.9 \\
            & A-MEM        & \underline{66.9} & 28.1 & 27.7 & 43.5 & 53.1 & 57.5 & 81.6 & \textbf{51.9} & 47.7 \\
        \midrule
        \multicolumn{11}{l}{\textbf{Visual Games}} \\
        \midrule
        \multirow{2}{*}{GPT-5.1}
            & In-context   & \textbf{63.0} & 36.5 & 17.1 & \underline{17.6} & \underline{43.7} & 23.5 & \textbf{74.7} & \textbf{43.8} & \textbf{39.0} \\
            & A-MEM        & \underline{57.5} & \textbf{45.1} & 12.4 & \textbf{24.3} & \textbf{46.5} & 19.3 & \underline{65.3} & \underline{42.1} & \underline{35.5} \\
        \cmidrule{1-11}
        \multirow{2}{*}{Qwen3-VL-32B-Instruct}
            & In-context   & 33.6 & \underline{43.2} & \underline{19.3} & 9.8 & 41.5 & 31.2 & 63.2 & 35.2 & 29.1 \\
            & A-MEM        & 37.4 & 39.1 & \textbf{20.2} & 11.2 & 38.2 & 34.4 & 64.4 & 36.3 & 31.2 \\
        \cmidrule{1-11}
        \multirow{2}{*}{InternVL3.5-38B}
            & In-context   & 34.0 & 35.6 & 13.6 & 6.1 & 33.3 & \underline{40.0} & 54.9 & 32.0 & 28.3 \\
            & A-MEM        & 37.2 & 33.1 & 13.9 & 9.5 & 35.3 & \textbf{44.7} & 61.3 & 34.9 & 30.7 \\
        \bottomrule
    \end{tabular}%
    }
    \caption{Main results under \textbf{self-experience evaluation} on text-only and visual games (ACC per ability; Overall reports ACC/F1). Within each modality block, the best and second-best numbers in each column are \textbf{bolded} and \underline{underlined}, respectively.}
    \label{tab:main-experiments}
\end{table*}

\section{Experiments}

\subsection{Model and Agent Baselines}
\label{sec:baselines}

We evaluate both text-only and vision-language settings. In each setting, we compare (i) foundation models used directly as reasoners (\emph{in-context}), and (ii) the same backbones equipped with explicit, persistent memory modules that store and retrieve information beyond the prompt window.

\paragraph{Language Models (Text-Only).}
We use a strong proprietary LLM (\textit{GPT-5.1}~\cite{gpt51}) and two open-weight long-context models, \textit{Qwen2.5-32B-Instruct}~\cite{qwen25} and \textit{Qwen3-32B}~\cite{qwen3}. The two successive Qwen generations at a similar 32B scale let us observe how backbone progression---spanning both architecture and training-recipe changes---shifts in-context memory ability.

\paragraph{Vision-Language Models.}
For visual games, we adopt \textit{GPT-5.1} and two open VLMs, \textit{Qwen3-VL-32B-Instruct}~\cite{qwen3vl} and \textit{InternVL3.5-38B}~\cite{internvl}, which jointly process game frames together with the associated textual logs (and HUD text when available). In both settings, the \emph{in-context} baseline answers questions by conditioning directly on the episode trajectory in the prompt, without any external memory state.

\paragraph{Memory Agents.}
On top of the above backbones, we evaluate representative memory agents that \emph{explicitly} maintain a persistent memory store and retrieve from it during question answering. In our main experiments, we focus on three complementary designs: \textbf{Mem0}~\cite{mem0}, which extracts and consolidates salient information into a reusable memory layer; \textbf{LangMem}~\cite{langmem}, which provides practical primitives for long-term memory extraction and adaptation in agent workflows; and \textbf{A-MEM}~\cite{amem}, an agentic note- and link-based memory system that organizes memories into an evolving network. For visual games, we pair backbones with A-MEM only: Mem0 and LangMem currently cannot ingest visual observations, so the visual comparison is between in-context prompting and A-MEM.

\begin{figure*}[t]
  \centering
  \includegraphics[width=\linewidth]{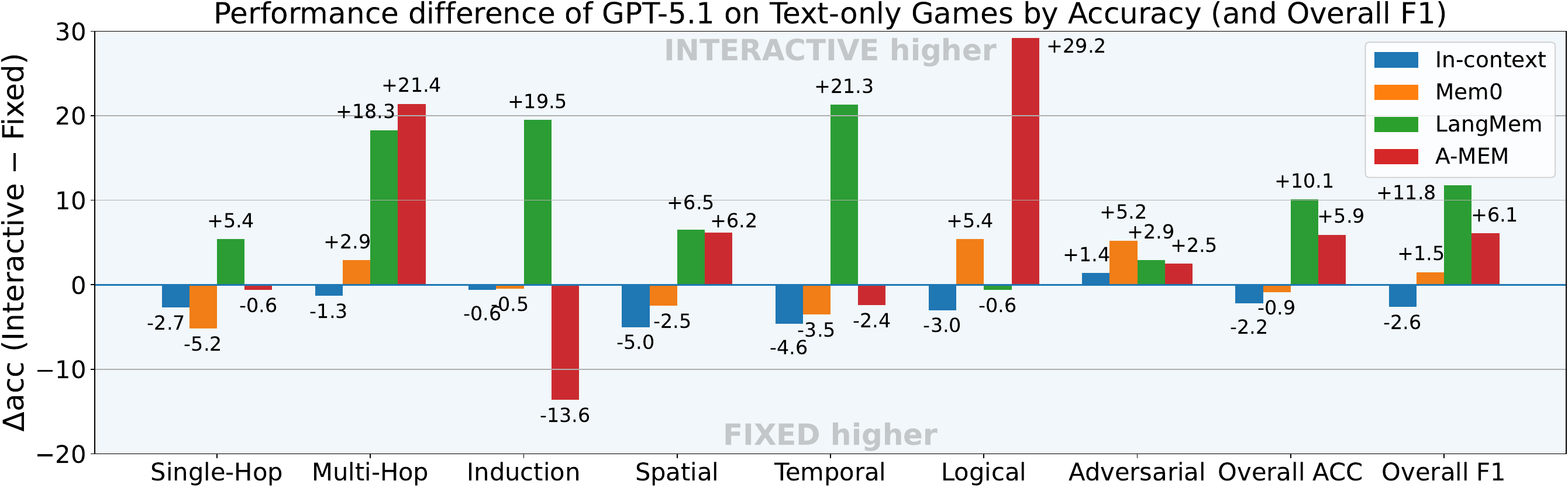}
  \caption{Performance gap between self-experience and controlled-exposure evaluation for GPT-5.1-based methods on text-only games. Each bar reports $\Delta\mathrm{acc}=\mathrm{acc}_{\text{interactive}}-\mathrm{acc}_{\text{fixed}}$ (percentage points) per question category; the last two columns show Overall Acc/F1. Positive values favor the interactive (self-experience) setting, negative values the fixed (controlled-exposure) setting.}
  \label{fig:interactive_and_fixed}
\end{figure*}

\subsection{Self-Experience Evaluation}
\label{sec:main_results_self_experience}
Table~\ref{tab:main-experiments} summarizes performance across ability types under \textit{self-experience evaluation}, where each method is evaluated on questions generated from its own interaction trajectory. Across models, Qwen3 achieves a higher score than Qwen2.5 in the in-context setting (Overall ACC: 40.8$\rightarrow$44.9), indicating stronger memory ability in the backbone itself. GPT-5.1 remains the strongest in-context model in both text-only games (49.7) and visual games (43.8). Among explicit memory agents, A-MEM achieves the best overall performance on text games (51.9 ACC, 47.7 F1).

\textbf{The benchmark remains challenging overall.} Even in the strongest settings, performance reaches only 51.9\% ACC on text games and 43.8\% on visual games, leaving substantial room for improvement in both memory and reasoning. The larger gap in the visual setting suggests that retaining and retrieving fine-grained, spatially grounded evidence under partial observability is still difficult: agents must bind transient observations to specific episodes and integrate them over time, and current memory modules often blur such visual details.
 
\textbf{Induction and spatial reasoning are the hardest abilities, especially in visual games.}
Induction peaks at only 34.2 on text-only games and 20.2 on visual games, while visual Spatial remains extremely low (best 24.3). This suggests that environment-grounded pattern discovery and spatial reasoning over remembered experience are the main bottlenecks for current memory agents. By contrast, memory agents perform relatively better on Single-Hop and Temporal questions.

\textbf{Explicit memory agents provide clear gains for open backbones on text-only games, but not universally.}
For Qwen2.5-32B and Qwen3-32B, A-MEM substantially improves Overall ACC over in-context prompting (40.8$\rightarrow$51.4 and 44.9$\rightarrow$51.9, respectively), and also improves F1. For GPT-5.1, however, plain in-context prompting remains competitive with the best memory agent (49.7 vs.\ 49.9 ACC, and the best F1 overall): a sufficiently strong long-context backbone can offset the benefit of external memory. On visual games, gains are smaller or sometimes negative, highlighting the need for stronger vision-oriented memory agents. Across ability types, LangMem and A-MEM often improve compositional abilities such as Spatial, Temporal, and Logical reasoning while sometimes reducing Single-Hop performance: explicit memory helps organize and retrieve longer-range evidence, but adds a retrieval bottleneck for facts a strong model could otherwise recall directly from context.

\subsection{Controlled-Exposure Evaluation}
\label{sec:controlled_exposure_eval}

To complement self-experience evaluation, we also report a \textit{controlled-exposure} setting: for each Jericho game, GPT-4.1~\cite{gpt41} plays one episode, whose log we freeze as a fixed interaction history, and all methods are evaluated on this same history and the same generator-derived questions. GPT-4.1 lies outside the evaluated model set, so no method answers questions over a history produced by its own backbone. The trajectory itself is not synthetic text: the model plays through the environment API, so every action is executed by the game engine and every observation, reward, and state transition is deterministically computed by the environment.

Figure~\ref{fig:interactive_and_fixed} shows the resulting performance gap, $\Delta \mathrm{acc}=\mathrm{acc}_{\text{interactive}}-\mathrm{acc}_{\text{fixed}}$. We find that the in-context baseline is systematically favored by the fixed-history setting, whereas memory agents, especially LangMem and A-MEM, benefit more under self-experience evaluation. Since the question generation procedure and seed are fixed, this gap is not simply due to question-generation randomness, but to what the two settings measure.

Under controlled exposure, all methods can reconstruct answers from the same fixed log, favoring prompt-only reconstruction by strong long-context backbones; under self-experience evaluation, the agent only observes a partial trajectory generated by its own actions, so explicit storing, searching, and organizing mechanisms operate closer to their intended use case. The larger gaps on Multi-Hop and Temporal questions are consistent with this difference.

\subsection{Human Evaluation}
\begin{table}[t]
  \centering
  \scriptsize
  \setlength{\tabcolsep}{3pt}
  \resizebox{\columnwidth}{!}{%
  \begin{tabular}{lccc ccc}
    \toprule
    & \multicolumn{3}{c}{Open-book} & \multicolumn{3}{c}{Closed-book} \\
    \cmidrule(lr){2-4}\cmidrule(lr){5-7}
    Type & Acc(\%) & F1(\%) & Time(s) & Acc(\%) & F1(\%) & Time(s) \\
    \midrule
    Single-Hop  & 80.6 & 81.5 & 20.3 & 14.1 & 13.9 &  9.2 \\
    Multi-Hop   & 59.4 & 59.8 & 27.7 &  9.8 & 10.0 & 11.4 \\
    Induction    & 53.9 & 54.8 & 23.7 & 19.9 & 20.2 & 10.3 \\
    Spatial     & 54.9 & 56.7 & 24.8 & 16.5 & 16.4 & 11.4 \\
    Temporal    & 76.0 & 76.7 & 24.2 & 49.1 & 51.4 & 11.0 \\
    Logical     & 47.4 & 48.8 & 27.1 & 24.2 & 24.8 & 14.6 \\
    Adversarial & 74.3 &  -   & 15.2 & 41.8 &  -   &  9.5 \\
    \midrule
    Overall     & 65.6 & 63.8 & 23.3 & 23.2 & 19.8 & 10.8 \\
    \bottomrule
  \end{tabular}%
  }
\caption{Human performance averaged over annotators per question type. Acc/F1 are in \%.}
\label{tab:human_per_type}
\end{table}

We use a human study to calibrate what the benchmark measures. Questions such as exact inventory counts at a given step exceed ordinary human memory by design, so the informative signal is the open-book vs.\ closed-book contrast: closed-book performance reflects the inherent difficulty of unaided episodic recall, while open-book performance isolates the difficulty of locating and integrating evidence when it is fully accessible---the same challenge faced by memory agents. Human open-book accuracy thus serves as a reference point: an agent scoring below it is likely bottlenecked by retrieval or reasoning rather than by ill-formed questions.

We recruit 11 annotators, all proficient in English: all participate in the text-game study, and 3 additionally annotate the visual-game study. For each game and seed, an annotator plays a full episode under fixed seed, after which QA is generated from the resulting trajectory as in agent evaluation. Annotators then answer the same QA set twice: (i) closed-book, from memory only, and (ii) open-book, with access to the same context file provided to the in-context model.

Table~\ref{tab:human_per_type} shows that open-book accuracy (65.6\%) far exceeds closed-book accuracy (23.2\%): many questions genuinely exceed human memory, yet become answerable once evidence is restored. Humans perform best on Single-Hop and Temporal questions, while Multi-Hop, Induction, Spatial, and Logical remain harder. The 65.6\% figure is a conservative ceiling estimate: multi-answer questions are credited for any acceptable answer via list-based scoring (Appendix~\ref{app:score_rules}), and part of the remaining loss stems from time-limited evidence search---relaxing the 60-second limit to 180 seconds raises open-book accuracy to 71.6\%. The visual-game study shows the same pattern at lower levels (open-book 59.2\% vs.\ closed-book 15.8\%; Table~\ref{tab:human_per_type_visual}), with more frequent skipping, reflecting the higher verification burden of visually grounded evidence. Appendix~\ref{app:humaneval} provides the full protocol and further analyses (open-book error types, time-limit sensitivity, non-response statistics, response-time buckets, and inter-annotator agreement).

%% file: 5_analysis.tex
\section{Further Analysis of Benchmark}

\subsection{Stability Across Random Seeds}\label{randomseed}
We investigate whether the evaluation results are stable across seeds, which affect text games via triggered random events and the visual game via map generation and entity spawning. We run multiple seeds and report the standard deviation (SD) of accuracy: on text games, $0.0364$ for GPT-5.1 and $0.0195$ for Qwen3-32B; on Crafter, $0.0453$ for GPT-5.1 and $0.0119$ for Qwen3-VL-32B. While some fluctuation is unavoidable due to in-game stochasticity, all SDs remain below $0.05$ ($\sim 0.02$ for Qwen3 models), suggesting limited seed-level variation across the evaluated runs.

\subsection{Correlation with other benchmarks}
\label{sec:cross_benchmark_corr}

Since \cref{sec:controlled_exposure_eval} shows that fixed-history evaluation can change the relative advantages of memory methods, we compare our findings with the fixed benchmarks \textbf{LongMemEval} and \textbf{LoCoMo}, using ability-level gains over each benchmark's own in-context baseline (Qwen-family backbones). The overall method ordering agrees: A-MEM is the strongest method on LongMemEval and in our Qwen3-32B text-game results. However, \emph{where} the gains come from differs sharply: on LoCoMo, external memory helps temporal questions (gains of 16.7 to 34.9 points), because they reduce to reconstructing cross-session timelines from retrievable dialogue facts; in our benchmark, the same methods \emph{hurt} temporal accuracy (drops of 7.1 to 13.9 points), because step-level ordering grounded in partially observed interaction is blurred by compressive memory and imperfect retrieval. This mismatch supports our claim that fixed-history QA and experience-conditioned QA measure different aspects of memory. Full comparison protocol and per-type gains are in Appendix~\ref{app:crossbench}.

\begin{figure}[t]
  \centering
  \includegraphics[width=0.75\linewidth]{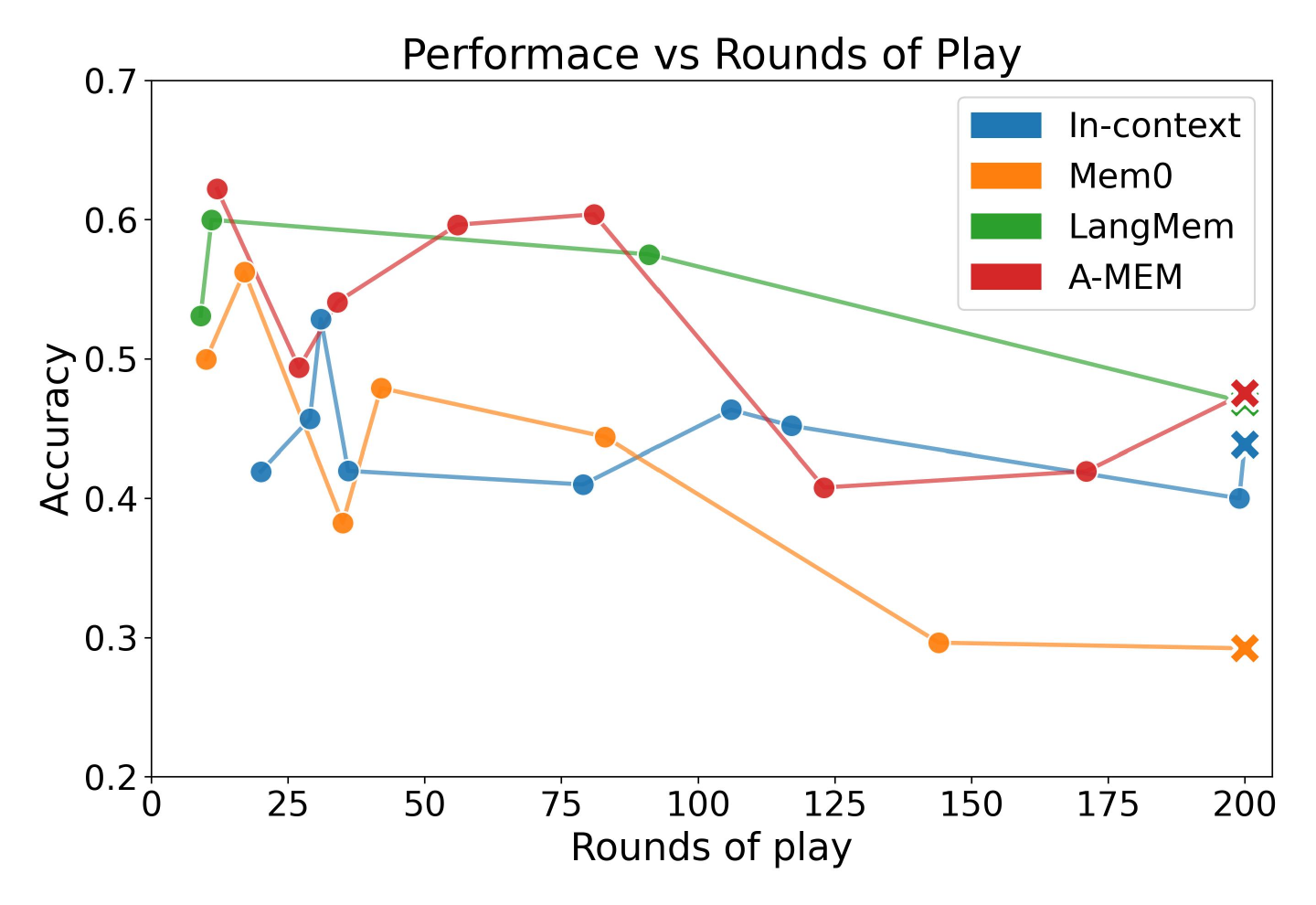}
  \caption{Performance vs.\ rounds of play. Each dot is one run, and each color is one method.}

  \label{fig:perf_vs_rounds}
\end{figure}

\subsection{Performance vs. Rounds of Play}\label{roundofplay}

We investigate how a run's overall accuracy varies with the number of interaction rounds during play. Figure~\ref{fig:perf_vs_rounds} plots each run as a point, summarizing runs that reach the 200-step cap by their mean accuracy. Runs with very few rounds are rare, and accuracies are generally (but not dramatically) higher below 100 rounds, suggesting performance is reasonably stable even under short runs.

Across methods, we still observe an overall negative relationship between accuracy and played rounds: for Mem0, mean accuracy drops from 0.444 (runs shorter than 200 steps) to 0.292 (capped-at-200 group; Pearson $r=-0.864$, $p=0.012$); A-MEM (0.526$\rightarrow$0.475) and LangMem (0.569$\rightarrow$0.469) show the same trend with smaller magnitudes. Longer runs expose more long-horizon memory demands, so comparing accuracies across unequal play budgets may not be strictly fair---this motivates our query-horizon control setting. Appendix~\ref{app:gameplay_coverage} further relates performance to game coverage.

\begin{figure}[t]
  \centering
  \includegraphics[width=0.85\linewidth]{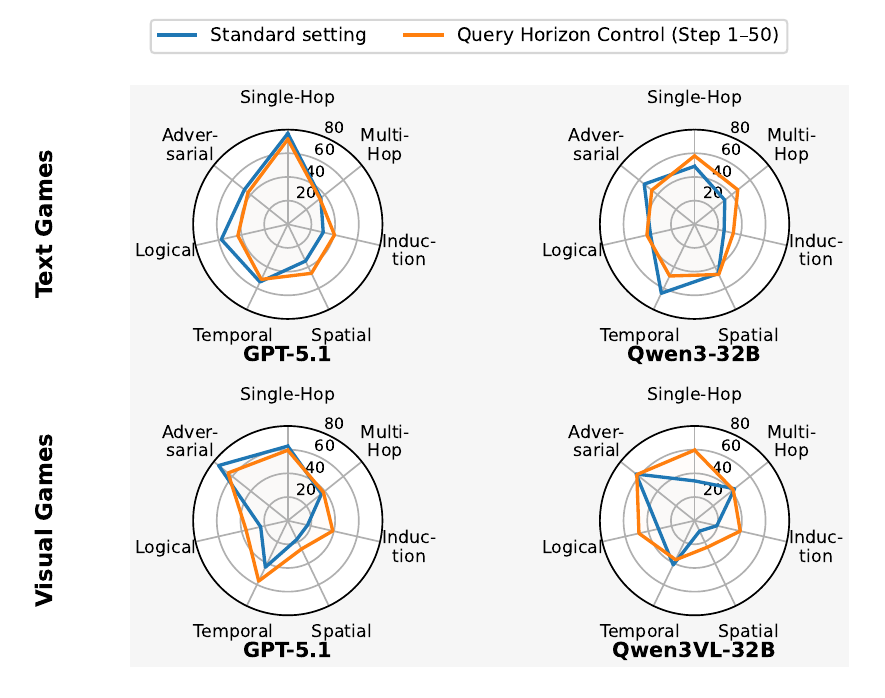}
  \caption{Query horizon control (steps 1--50) under the in-context baseline: standard vs.\ horizon-controlled setting across seven skill categories.}
  \label{fig:query_horizon_radar}
\end{figure}

\subsection{Query Horizon Control Setting}\label{queryhorizon}

\Cref{fig:query_horizon_radar} shows that horizon control changes the benchmark's difficulty profile rather than uniformly making it easier. In visual games, \textbf{induction} improves substantially, consistent with long-range pattern questions becoming more local when the queried window is shortened, while \textbf{Adversarial} and some multi-evidence categories can degrade, suggesting the added scope constraint increases the burden of retrieving within-window evidence. The setting thus reshapes error types and offers a complementary view of episodic memory beyond the standard evaluation; full results are in \cref{app:queryhorizon}.

\subsection{Failure Mode Analysis}
\label{sec:failure_modes}

To understand \emph{why} current systems fall short, we manually inspect agent errors and identify five recurring failure modes (case studies in Appendix~\ref{app:case_studies}): (1) \textbf{evidence localization errors}---the agent misses the anchor step of a multi-hop query and falsely declares it not answerable; (2) \textbf{long-horizon aggregation failures}---set- or count-style questions accumulate silent omissions and deduplication mistakes, producing systematic undercounts; (3) \textbf{visual grounding mistakes}---nearby salient content is confused with the precisely queried target (e.g., adjacent sand vs.\ the tile under the player); (4) \textbf{reasoning confusion}---attempts are conflated with verified outcomes (e.g., counting chop actions instead of successful wood collections); and (5) \textbf{answer--explanation inconsistency}---the rationale derives the correct value but the emitted answer differs. Two question families remain unsolved---global object counting under partial observability and obstacle-aware navigation---both requiring integrating observations into a coherent global map over time.

Across the inspected cases, the errors arise in retrieval, aggregation, grounding, or answer finalization rather than in question construction, and human open-book errors show parallel patterns (Appendix~\ref{app:humaneval}). Localization and aggregation errors dominate the inspected cases, consistent with induction and spatial remaining the bottlenecks in Table~\ref{tab:main-experiments}: both compose many small pieces of evidence, so a single missed anchor or omitted instance corrupts the final answer.

%% file: 6_conclusion.tex
\section{Conclusion}
We presented a game-based, trajectory-dependent benchmark generator for long-term memory evaluation in text and vision-language settings. By generating QA from an agent's own experience with deterministic ground truth and controlled answerability, \benchname supports fine-grained diagnosis of memory skills beyond static long-context QA, complemented by a controlled-exposure setting. Current systems remain far from saturated: induction and spatial reasoning are the primary bottlenecks, and persistent memory yields meaningful gains, most clearly on open backbones.

%% file: 7_appendix.tex
\input{9_appendix_coverage}
\input{9_appendix_human}

\section{Experiment Details}\label{app:expdetails}

\subsection{In-context Prompting}
A common way to equip LLMs with ``memory'' is to treat it as external context. Two dominant approaches are helpful: (i) \textbf{long-context (LC)} modeling, which extends the context window to include long histories, and (ii) \textbf{retrieval-augmented generation (RAG)}, which retrieves relevant snippets from external context. Both enable question answering over large information sources, but they differ in what they optimize: LC emphasizes end-to-end reasoning over a long input, while RAG emphasizes \emph{selecting} a short, relevant subset before reasoning ~\cite{lc_vs_rag}.

However, both LC and RAG are typically evaluated as document-understanding QA: the ``memory'' is assumed to be a read-only record, and the core challenge is to locate evidence and generate an answer~\cite{ragfornlptasks}. This differs from episodic memory in interactive settings, where an agent decide what to store from its own experience, and when to revise it. In addition, even with a long context window, the model can still ignore key memory from the middle of a long prompt or when mixed with noisy memory~\cite{lostin}. RAG has an explicit bottleneck---retrieval accuracy~\cite{dpr}---where errors in indexing, querying, or ranking can hide crucial evidence. These gaps motivate memory agents that explicitly manage a persistent store, rather than only expanding or retrieving context at inference time~\cite{survey_agent}.
\FloatBarrier

\subsection{Memory Agents}
Memory agents vary mainly in memory management mechanism---from plug-and-play memory layers, to structured and self-organizing memory networks, to learned policies that optimize memory operations from feedback~\cite{survey_memory}. Below we group representative methods into: 1) heuristic memory layers, 2) agentic memory structures, and 3) policy-learned memory.

\paragraph{Heuristic Memory Layers}
A practical line of work builds ``memory layers'' that extract facts from interactions, consolidate redundancies, and retrieve relevant memories at inference time. Mem0~\cite{mem0} emphasizes scalable long-term memory with explicit and efficient retrieval, including using graph based representations for relational information. LangMem~\cite{langmem} provides a developer-facing toolkit that focuses on reusable primitives for memory retrieval, and connects memory to agent behavior updates with integration in agent frameworks. These systems are easy plug-ins, but their write behavior is often affected by prompts~\cite{memgpt}.

\paragraph{Agentic Memory Structures}
Another direction treats memory management as an explicit system problem. MemGPT~\cite{memgpt} proposes an OS-inspired hierarchy where the agent manages multiple memory tiers and performs context ``swaps'' so that a limited-context model can behave as if it had a much larger working memory. Going further on organization, A-MEM~\cite{amem} builds an interconnected memory network using structured notes, link generation, and memory evolution so new experiences can refine existing memories over time. These designs aim to organize memories into structured and evolving networks, so evaluation depends on whether the benchmark probes memory organization and updates~\cite{MemoryAgentBench}, not only retrieval accuracy.

\paragraph{Policy-Learned Memory}
Recent work starts to learn memory operations directly with reinforcement learning~\cite{mema}. MemAgent~\cite{memagent} trains a memory agent for long-context processing, reading inputs in segments and updating a fixed-size memory state to support extremely long contexts. Memory-R1~\cite{memory-r1} explicitly parameterizes memory operations and uses outcome-driven RL to learn when to modify memory and how to use retrieved entries for answering. 

\subsection{Prompts}
\begin{table*}
  \centering
  \scriptsize
  \setlength{\tabcolsep}{4pt}
  \renewcommand{\arraystretch}{1.12}
  \begin{tabular}{p{0.22\textwidth} p{0.74\textwidth}}
    \toprule
    \textbf{Component} & \textbf{Prompt text (template)} \\
    \midrule
    Icon legend message &
    Crafter: ``ICON LEGEND: reference image for object recognition (Player, Plant, Cow, Zombie, Skeleton, Arrow, Water, Sand, Grass, Tree, Path, Stone, Coal, Iron, Diamond, Lava, Table, Furnace). Use this as ground truth for appearance.''  \\
    \midrule
    QA Paraphrase &
    You are a precise paraphraser for some evaluation questions. Your goal is to improve sentence style variation. Strict rules: 1) Do NOT change meaning. 2) Do NOT change any numbers or step indices. 3) Preserve domain terms (actions/resources/terrains) as-is. 4) English only. Output a single sentence without extra commentary. Paraphrase the following sentence without changing its meaning or numbers:<{text}>  \\
    \midrule
    \multicolumn{2}{l}{\textbf{Agent Game Playing Prompts}} \\
    \midrule
    Jericho system prompt &
    All content in this prompt is purely fictional, coming from a text-based game log. \newline
    It does not describe or endorse real-world harmful behavior, and should be treated only as harmless game simulation. \newline
    You are an expert player of classic parser interactive fiction game. \newline
    You will receive the current observation and optional candidate actions. \newline
    Your task each turn: decide ONE valid parser command. \newline
    Rules: \newline
    - Output STRICT JSON only (no code fences, no extra text). \newline
    - JSON schema: \texttt{\{"action": "\textless{}string\textgreater{}", "reason": "\textless{}string\textgreater{}"\}}. \newline
    - ``action'' must be a single parser command (lowercase, concise), e.g., ``open mailbox'', ``take leaflet'', ``north'', ``look''. \newline
    - Read carefully from history environments and actions to help you make decisions. \newline
    - From history, avoid taking repetitive or loop actions, like keep taking then putting down an item, or looking environments that you have looked. \newline
    - Avoid meta-commands like save/restore/quit. Do not include punctuation beyond the command itself. \newline
    - Consider candidate\_actions. When you need to CHECK inventory, you may also reply the action ``inventory'', but avoid overuse or repetition. \\

    \midrule
    Jericho per-turn user template &
    Current step: \texttt{\{step\_index\}} \newline
    Current observation: \newline
    \texttt{\{observation\}} \newline
    Score: \texttt{\{score\}/\{max\_score\}} \textbar\ Moves: \texttt{\{moves\}} \newline
    Location: \texttt{\{location\}} \newline
    Candidate actions (not exhaustive): \texttt{\{valid\_actions\}}. \newline\newline
    Context of the last \texttt{\{history\_turns\}} turns (ONLY observation and the agent action per turn, most recent last): \newline
    \texttt{\{history\_snippet\}} \newline
    Respond with STRICT JSON only: \newline
    \texttt{\{"action":"\textless{}one command\textgreater{}", "reason":"\textless{}one short sentence\textgreater{}"\}} \\

    \midrule
    Crafter system prompt &
    You are a vision game agent for the Crafter environment.
    You will see a single 64x64 RGB frame and a HUD(text) line with the last step's numeric stats.
    Choose EXACTLY ONE action and respond with STRICT JSON only: \newline
    \texttt{\{"action\_id": \textless{}0-16\textgreater{}, "action\_name": "\textless{}string\textgreater{}", "reason": "\textless{}brief diagnosis \& plan (no raw numbers)\textgreater{}"\}} \newline
    Valid actions: \texttt{0:NOOP, 1:MOVE\_LEFT, 2:MOVE\_RIGHT, 3:MOVE\_UP, 4:MOVE\_DOWN, 5:DO, 6:SLEEP, 7:PLACE\_STONE, 8:PLACE\_TABLE, 9:PLACE\_FURNACE, 10:PLACE\_PLANT, 11:MAKE\_WOOD\_PICKAXE, 12:MAKE\_STONE\_PICKAXE, 13:MAKE\_IRON\_PICKAXE, 14:MAKE\_WOOD\_SWORD, 15:MAKE\_STONE\_SWORD, 16:MAKE\_IRON\_SWORD.} \newline
    (Runtime appends: ``Game rules: \texttt{\{instructions\}}''.) \\

    \midrule
    Crafter per-step user template &
    \texttt{\{hud\_text\}} \newline
    Env: \texttt{\{env\_name\}}. Step \texttt{\{display\_step\}}. Recent actions: \texttt{\{recent\}}.
    Use HUD(text) as ground truth; DO NOT restate raw numbers in your reason.
    An ICON LEGEND image is provided once in this request for recognition; match tiles/objects to it.
    Provide a brief diagnosis (e.g., thirsty/hungry/sleepy/safe) and the plan. \\

    \midrule
    \multicolumn{2}{l}{\textbf{Agent QA answering Prompts}} \\
    \midrule
    Jericho QA system prompt &
    You answer Jericho QA concisely. Reply with JSON only. \\

    \midrule
    Jericho QA batch header &
    You will answer multiple questions about the same episode. \newline
    Return strictly JSON. \newline
    For a single question: \texttt{\{"answer": "\textless{}short\textgreater{}", "explanation": "\textless{}brief\textgreater{}"\}} \newline
    For batched questions: an array of objects, each:
    \texttt{\{"id": "\textless{}qid\textgreater{}", "answer": "\textless{}short\textgreater{}", "explanation": "\textless{}brief\textgreater{}"\}} \newline
    Keep explanation short. \newline
    If your answer is a step number, answer only the number. \newline
    For any question you find not answerable, strictly reply ``not answerable'' as answer, and give explanation. \newline
    Questions: \\

    \midrule
    Crafter QA system prompt &
    You are a concise answerer for questions about the Crafter environment.
    All provided steps are part of a single continuous episode of the game.
    The timeline presents actions, reasons, and health related stats for each step in order.
    Use the entire history to answer questions; recall what was seen or collected earlier even if it is not visible in the current frame.
    If there is not enough information in the given steps to answer, reply with ``not answerable'' as the answer and briefly explain why.
    Always reply in strict JSON format and do not guess. \newline
    (Runtime may append: ``Game rules: \texttt{\{instructions\}}''.) \\

    \midrule
    Crafter QA schema (used for batching) &
    Return strictly JSON. \newline
    For a single question: \texttt{\{"answer": "\textless{}short\textgreater{}", "explanation": "\textless{}brief\textgreater{}"\}} \newline
    For batched questions: an array of objects, each:
    \texttt{\{"id": "\textless{}qid\textgreater{}", "answer": "\textless{}short\textgreater{}", "explanation": "\textless{}brief\textgreater{}"\}} \newline
    Keep explanation short.
    For any question you find not answerable, strictly reply ``not answerable'' as answer, and give explanation. \newline
    (Batch header prepends: ``You will answer multiple questions about the same episode.'' and ``Questions:''.) \\

    \bottomrule
  \end{tabular}
\caption{Prompt templates used in our pipelines (gameplay and QA answering) for Jericho text games and the Crafter visual game.}
\label{tab:prompt_templates}
\end{table*}

We provide the prompts used in the pipeline in Table~\ref{tab:prompt_templates}.

\subsection{End-to-end Generation Procedure.}
Algorithm~\ref{alg:generator} summarizes the generator. It is deterministic given a trace $\tau$, game state $\mathcal{S}$, and a random seed used only for sampling among multiple valid template instantiations.
\begin{algorithm}
\caption{Trajectory-conditioned benchmark generation}
\label{alg:generator}
\begin{algorithmic}[1]
\Require Trace $\tau$, optional auxiliary state $\mathcal{S}$, template set $\mathcal{T}$, sampling seed $u$
\Ensure QA set $\mathcal{Q}$
\State Build timeline index $\mathcal{I} \gets \textsc{IndexTrace}(\tau,\mathcal{S})$
\State Build event index $\mathcal{E} \gets \textsc{ExtractEvents}(\mathcal{I})$
\State $\mathcal{Q} \gets \emptyset$
\ForAll{template $T \in \mathcal{T}$}
  \State $\mathcal{C} \gets T.\textsc{EnumerateCandidates}(\mathcal{I},\mathcal{E})$
  \State $\mathcal{C}' \gets \textsc{FilterByPreconditions}(\mathcal{C})$
  \State Sample a subset $\widehat{\mathcal{C}}$ from $\mathcal{C}'$ using seed $u$
  \ForAll{candidate $c \in \widehat{\mathcal{C}}$}
    \State $(q,y,m) \gets T.\textsc{Instantiate}(c,\mathcal{I},\mathcal{E})$
    \If{\textsc{ValidateInstance}$(q,y,m,\mathcal{I})$}
      \State $\mathcal{Q} \gets \mathcal{Q} \cup \{(q,y,m)\}$
    \EndIf
  \EndFor
\EndFor
\State \Return $\mathcal{Q}$
\end{algorithmic}
\end{algorithm}

\subsection{Rules for Score Calculation}
\label{app:score_rules}

We evaluate model outputs by scoring the extracted predictions against the reference answers. Our scorer is rule-based and selects different matching strategies according to the \texttt{answer\_type}. We normalize both prediction and reference by lowercasing, trimming whitespace, removing parenthesized spans, and stripping surrounding quotes.

\textbf{String.}
We first decide whether an answer should be \emph{exact-match} using simple patterns, including URLs, filenames, page-like identifiers, phone-number formats, time expressions containing ``a.m.''/``p.m.'', dates (YYYY-MM-DD / YYYY-MM), and email addresses. If exact-match is required, we perform direct string comparison and assign a score in $\{0,1\}$. Otherwise, we compute ANLS (Average Normalized Levenshtein Similarity) with a threshold $\tau = 0.5$ \cite{biten2019stvqa}:
\begin{equation}
\begin{split}
s(g,p) &= 1-\frac{d_{\text{lev}}(g,p)}{\max(|g|,|p|)} \\
\text{ANLS}(g,p) &=
\begin{cases}
s(g,p) & \text{if } s(g,p)>\tau,\\
0      & \text{otherwise.}
\end{cases}
\end{split}
\end{equation}

where $d_{\text{lev}}(\cdot,\cdot)$ is the Levenshtein edit distance.

\textbf{Integer.}
We cast both reference and prediction into an integer-like form, accepting common variants such as ``1'', ``1.0'', or strings with extra spaces (and optional trailing ``\%''). The score is $1$ iff the parsed integers are equal, otherwise $0$.

\textbf{Float.}
We treat prediction and reference as equal if they match after rounding with an adaptive precision (at least two decimal digits), or if they are within a $1\%$ relative tolerance. To reduce ambiguity between fractions and percentages, we also accept matches against $\{g/100,\, g,\, 100g\}$.

\textbf{List.}
Some questions admit multiple acceptable answers. We represent such references as a list of candidates.
We score the prediction against each candidate (using the same atomic rules for string/integer/float),
and take the maximum score as the final score for the example, i.e., a prediction is correct if it matches any one of the candidates \cite{rajpurkar2016squad}.

\textbf{Unanswerable and aggregate metrics.}
We use the canonical label \texttt{not answerable} for unanswerable questions \cite{vanlandeghem2023dude, mmlongbenchdoc}. Overall \textbf{Accuracy} is the mean of per-question scores. We additionally compute an \textbf{F1}-style score by treating \texttt{not answerable} as the negative class: recall is computed over questions with answer $\neq$ \texttt{not answerable}, precision is computed over questions with prediction $\neq$ \texttt{not answerable}, and F1 is their harmonic mean.
\FloatBarrier
\subsection{Model Hyper-Parameters}
Unless otherwise stated, all models use deterministic decoding with temperature $=0.0$ for both gameplay and question answering. We report results under two \emph{Query Horizon Control} settings: \texttt{QHC=-1} (no restriction) and \texttt{QHC=50} (all generated queries are constrained to steps 1--50). For QA construction, we use a fixed automatic generator with \texttt{max\_per\_type=2} and enable paraphrasing (paraphrase=True); in all experiments we evaluate on the paraphrased questions (source=paraphrase).

For text-only games, we cap each episode at 200 interaction steps. For the \emph{in-context} baseline, we include the most recent 10 turns of trajectory context (\texttt{history\_turns=10}). For memory-based agents, gameplay-time retrieval uses \texttt{top\_k=10}. During QA answering, we use a maximum output length of 1024 tokens; for memory-based agents, per-question retrieval uses \texttt{top\_k=8}. 

For visual game, we cap each episode at 200 interaction steps. For the \emph{in-context} baseline, we include the most recent 5 turns (\texttt{history\_turns=5}); for memory-based agents, gameplay-time retrieval uses \texttt{top\_k=5}. During gameplay we use \texttt{max\_tokens=128} for action generation, while QA answering uses \texttt{max\_tokens=4096} with batch size 4. When providing multiple frames to VLMs, we use a fixed mosaic configuration: \texttt{frames\_mode=mosaic}, \texttt{mosaic\_cols=10}, \texttt{mosaic\_cell=160}, \texttt{mosaic\_per\_image=200}, and \texttt{mosaic\_per\_batch=10}.
\FloatBarrier

\subsection{Compute and GPU Hours}
Across experiments, we measure approximate end-to-end cost/runtime per baseline (one backbone × one method, covering play → QA generation → answering). Using the OpenAI API with GPT-5.1, each text-game baseline costs about 15 USD and takes roughly 15 hours wall-clock, while each visual-game baseline costs about 5 USD and takes about 5 hours; these costs follow token-based pricing. For self-hosted open models, text-only runs on 2× NVIDIA H20-e, where Qwen3-32B takes \~10 hours/baseline (\~20 GPU-hours) and Qwen2.5-32B-Instruct takes ~6 hours/baseline (\~12 GPU-hours). The visual-game runs use 4× NVIDIA H800 (80GB), where Qwen3-VL-32B-Instruct and InternVL3.5-38B each take \~4 hours/baseline (\~16 GPU-hours).
\FloatBarrier

\FloatBarrier

\input{9_appendix_crossbench}

\input{9_appendix_horizoncontrol}

\input{9_appendix_casestudy}

\input{9_appendix_jericho}

\input{9_appendix_crafter}

\input{9_appendix_template}

\input{9_appendix_algorithm}

%% file: 9_appendix_coverage.tex
\section{Gameplay Performance and Exploration Coverage}
\label{app:gameplay_coverage}

\begin{table*}[t]
  \centering
  \scriptsize
  \setlength{\tabcolsep}{5pt}
  \begin{tabular}{llcccccc}
    \toprule
    Model & Seed & Ach & Tiles & Perf & Coverage & QA-Acc & QA-F1 \\
    \midrule
    GPT-5.1        & 1   & 6 & 94 & 27.3 & 22.9 & 38.3 & 32.9 \\
    GPT-5.1        & 42  & 5 & 75 & 22.7 & 18.3 & 42.0 & 38.0 \\
    GPT-5.1        & 43  & 9 & 54 & 40.9 & 13.2 & 50.8 & 47.1 \\
    GPT-5.1        & 100 & 7 & 34 & 31.8 & 8.3  & 43.5 & 40.1 \\
    GPT-5.1        & 123 & 8 & 83 & 36.4 & 20.3 & 43.9 & 36.4 \\
    \midrule
    Qwen3-VL-32B   & 1   & 9 & 38 & 40.9 & 9.3  & 34.1 & 30.9 \\
    Qwen3-VL-32B   & 42  & 2 & 33 & 9.1  & 8.1  & 33.4 & 25.7 \\
    Qwen3-VL-32B   & 43  & 4 & 61 & 18.2 & 14.9 & 35.8 & 27.8 \\
    Qwen3-VL-32B   & 100 & 4 & 37 & 18.2 & 9.0  & 35.5 & 31.8 \\
    Qwen3-VL-32B   & 123 & 2 & 24 & 9.1  & 5.9  & 37.4 & 29.8 \\
    \bottomrule
  \end{tabular}
  \caption{Full per-run data for Crafter. ``Perf'' denotes gameplay performance (achievements unlocked / 22), ``Coverage'' denotes visited tiles / 4096, and QA-Acc/QA-F1 denote memory QA performance on the same runs. All rates are shown as percentages.}
  \label{tab:coverage_full_crafter}
\end{table*}

\begin{table*}[t]
  \centering
  \scriptsize
  \setlength{\tabcolsep}{4pt}
  \begin{tabular}{llcccccccc}
    \toprule
    Model & Game & Score & Max Score & Visited Locs & Total Locs & Perf & Coverage & QA-Acc & QA-F1 \\
    \midrule
    GPT-5.1 & advent   & 82  & 350 & 33 & 146 & 23.4  & 22.6 & 46.3 & 43.1 \\
    GPT-5.1 & awaken   & 0   & 50  & 6  & 46  & 0.0   & 13.0 & 55.8 & 53.7 \\
    GPT-5.1 & balances & 10  & 51  & 8  & 43  & 19.6  & 18.6 & 45.2 & 40.6 \\
    GPT-5.1 & dragon   & -49 & 25  & 25 & 55  & -196.0& 45.5 & 44.9 & 41.6 \\
    GPT-5.1 & gold     & 3   & 100 & 10 & 94  & 3.0   & 10.6 & 43.0 & 37.5 \\
    GPT-5.1 & jewel    & 7   & 90  & 17 & 97  & 7.8   & 17.5 & 45.6 & 42.6 \\
    GPT-5.1 & karn     & 5   & 170 & 14 & 248 & 2.9   & 5.6  & 40.7 & 39.4 \\
    GPT-5.1 & ludicorp & 14  & 150 & 14 & 50  & 9.3   & 28.0 & 40.0 & 38.3 \\
    GPT-5.1 & moonlit  & 0   & 1   & 7  & 21  & 0.0   & 33.3 & 34.2 & 35.0 \\
    GPT-5.1 & pentari  & 5   & 70  & 11 & 33  & 7.1   & 33.3 & 41.9 & 39.9 \\
    GPT-5.1 & reverb   & 0   & 50  & 9  & 63  & 0.0   & 14.3 & 52.9 & 50.4 \\
    GPT-5.1 & sorcerer & 20  & 400 & 9  & 61  & 5.0   & 14.8 & 41.0 & 39.4 \\
    GPT-5.1 & zork1    & 30  & 350 & 11 & 107 & 8.6   & 10.3 & 42.0 & 39.7 \\
    GPT-5.1 & zork2    & 5   & 400 & 16 & 19  & 1.2   & 84.2 & 42.8 & 43.4 \\
    GPT-5.1 & zork3    & 3   & 7   & 9  & 12  & 42.9  & 75.0 & 45.7 & 40.9 \\
    \midrule
    Qwen2.5-32B & advent   & 36   & 350 & 11 & 146 & 10.3   & 7.5  & 33.7 & 29.7 \\
    Qwen2.5-32B & awaken   & 0    & 50  & 6  & 46  & 0.0    & 13.0 & 30.6 & 32.8 \\
    Qwen2.5-32B & balances & 10   & 51  & 3  & 43  & 19.6   & 7.0  & 46.6 & 47.2 \\
    Qwen2.5-32B & dragon   & -107 & 25  & 15 & 55  & -428.0 & 27.3 & 26.1 & 23.2 \\
    Qwen2.5-32B & gold     & 3    & 100 & 9  & 94  & 3.0    & 9.6  & 39.1 & 38.0 \\
    Qwen2.5-32B & jewel    & 2    & 90  & 14 & 97  & 2.2    & 14.4 & 29.9 & 29.8 \\
    Qwen2.5-32B & karn     & 0    & 170 & 13 & 248 & 0.0    & 5.2  & 27.1 & 22.2 \\
    Qwen2.5-32B & ludicorp & 8    & 150 & 8  & 50  & 5.3    & 16.0 & 46.5 & 43.8 \\
    Qwen2.5-32B & moonlit  & 0    & 1   & 5  & 21  & 0.0    & 23.8 & 29.6 & 27.2 \\
    Qwen2.5-32B & pentari  & 5    & 70  & 16 & 33  & 7.1    & 48.5 & 39.0 & 37.6 \\
    Qwen2.5-32B & reverb   & 0    & 50  & 11 & 63  & 0.0    & 17.5 & 37.1 & 32.4 \\
    Qwen2.5-32B & sorcerer & 5    & 400 & 7  & 61  & 1.2    & 11.5 & 41.5 & 42.6 \\
    Qwen2.5-32B & zork1    & 30   & 350 & 13 & 107 & 8.6    & 12.1 & 37.9 & 36.1 \\
    Qwen2.5-32B & zork2    & -10  & 400 & 6  & 19  & -2.5   & 31.6 & 41.3 & 41.1 \\
    Qwen2.5-32B & zork3    & 3    & 7   & 9  & 12  & 42.9   & 75.0 & 44.1 & 43.5 \\
    \bottomrule
  \end{tabular}
  \caption{Full per-game data for Jericho. ``Perf'' denotes normalized in-game score (\texttt{score}/\texttt{max\_score}), ``Coverage'' denotes visited locations / total locations, and QA-Acc/QA-F1 denote memory QA performance on the same runs. All rates are shown as percentages. Note that ``Perf'' can be negative in rare penalty cases.}
  \label{tab:coverage_full_jericho}
\end{table*}

A natural concern for trajectory-conditioned evaluation is whether memory QA performance is largely driven by how well an agent plays or how much of the environment it explores. To clarify this, we analyze two trajectory-level quantities for the runs used to generate QA.

For \textbf{Crafter}, we measure \textit{gameplay performance} as the number of achievements unlocked divided by 22, and \textit{exploration coverage} as the fraction of distinct visited tiles in the $64 \times 64$ world. For \textbf{Jericho}, we measure \textit{gameplay performance} as normalized in-game score ($\texttt{score}/\texttt{max\_score}$) and \textit{exploration coverage} as the fraction of visited locations over all locations parsed from the game ROM. For Jericho, rare negative normalized scores are excluded from correlation analysis because they can disproportionately distort summary statistics.

Overall, we find that \textbf{exploration coverage is not a consistent predictor of memory QA accuracy}. Across both Crafter and Jericho, the correlations between coverage and QA accuracy are weak and unstable, sometimes near zero or even negative. This suggests that \benchname is not simply rewarding agents for visiting more places, and that broader coverage alone does not deterministically make the resulting QA either easier or harder.

By contrast, \textbf{gameplay performance can matter in some settings, but its effect is environment- and model-dependent}. In Crafter with GPT-5.1, achievement performance shows a strong positive association with QA accuracy, while coverage does not. A plausible explanation is that achievements reflect more meaningful event and state transitions, which create richer episodes for episodic recall. In Jericho with Qwen2.5-32B, gameplay performance also shows a moderate positive association with QA accuracy, whereas coverage remains weak. This suggests that more successful task progress can produce trajectories with denser and more coherent event chains, but simply visiting more locations is not the dominant factor.

These results clarify the scope of \benchname. Gameplay proficiency can influence the informational richness of the collected experience, but the benchmark is not reducible to a gameplay or exploration leaderboard. In particular, the weak and inconsistent relationship between coverage and QA accuracy supports our goal of measuring episodic memory conditioned on the agent's own experience, rather than merely measuring who explores more. Detailed summary statistics and full correlation tables are provided in Table \ref{tab:coverage_full_crafter} and \ref{tab:coverage_full_jericho}.

%% file: 9_appendix_human.tex
\section{Human Studies Details and In-depth Analysis}
\label{app:humaneval}

This appendix provides full details of our human evaluation protocol and additional analyses used to interpret human performance, including open-book error types, time-limit sensitivity, non-response and abstention behavior, and accuracy by response-time bucket.

\begin{table*}[t]
  \centering
  \scriptsize
  \setlength{\tabcolsep}{5pt}
  \renewcommand{\arraystretch}{1.15}
  \begin{tabular}{p{0.18\textwidth} p{0.77\textwidth}}
    \toprule
    \textbf{Stage} & \textbf{What to do (key rules only)} \\
    \midrule
    Deliverables &
    Submit a single zip containing: (i) your \textbf{gameplay log} (auto-generated while you play) and
    (ii) your \textbf{answers} for \textbf{both} rounds (close-book and open-book), including the generated QA files kept alongside the answer file. \\
    \midrule
    1) Play the game &
    \textbf{Goal:} play as well as you can based on your understanding. \newline
    \textbf{How to play:} read the \emph{Observation} carefully each step, then choose an action; explore the map; avoid random actions. \newline
    \textbf{If stuck:} if you make no progress (looping/confused), you may quit after roughly $\sim$100 steps. \\
    \midrule
    GUI (during play) &
    The interface typically shows: \newline
    \textbf{Left:} status (score/moves/location) and the current \emph{Observation} text. \newline
    \textbf{Right:} available actions as buttons, plus an input box for a custom action; you can also open \textbf{inventory}. \\
    \midrule
    2) Generate questions (QA) &
    After play, run the provided pipeline to generate questions from \textbf{your own log}. \newline
    You only need to know: the game name and your run folder / log identifier. \\
    \midrule
    3) Answer questions (two rounds) &
    You must complete the same question set twice: \newline
    \textbf{Round A: Closed-book.} Answer only from memory; do \textbf{not} look back at any logs/records. If unsure, use \textbf{not answerable}. \newline
    \textbf{Round B: Open-book.} You may consult the provided \textbf{QA context file} (generated from your log) and answer again. \\
    \midrule
    GUI (during answering) &
    The answering window provides: \newline
    A question panel with progress and a per-question \textbf{time limit} (auto-advances when time is up). \newline
    A text box to enter your answer and confirm. \newline
    Buttons: \textbf{not answerable} (cannot be answered even with available info) and \textbf{cannot remember} (you believe it happened but you cannot recall). \newline
    You can \textbf{save and exit} and later continue from where you stopped. \\
    \bottomrule
  \end{tabular}
\caption{Human evaluation instructions for Jericho text games (play $\rightarrow$ auto-QA $\rightarrow$ answer twice).}
\label{tab:jericho_human_eval_instructions}
\end{table*}

\paragraph{Annotators.}
We recruited 11 annotators, all proficient in English. All 11 participated in the Jericho text-game study, and 3 of them additionally participated in the Crafter visual-game study. Consent was obtained for the usage of evaluated results. The data collection protocol was approved by an ethics review board. The instruction sheet used in the Jericho human study is shown in Table~\ref{tab:jericho_human_eval_instructions}.

\paragraph{Task and QA construction.}
For each game and seed, an annotator plays a full episode under a fixed seed while we log the interaction trajectory. We then generate QA from the resulting trajectory using the same dynamic procedure as in our agent evaluation, except that we exclude templates that depend on explicit agent reasoning because they are not applicable to the human playing process.

\paragraph{Answering protocol (closed-book vs. open-book).}
Annotators answer the same QA set twice: (i) closed-book, answering from memory only without consulting any records; and (ii) open-book, answering with access to the trajectory context file (consistent with the agent-evaluation setting and not containing any hidden backend information). For visual games, open-book additionally allows viewing the saved frames, again consistent with the agent-evaluation setting.

\begin{figure}[t]
  \centering
  \includegraphics[width=\linewidth]{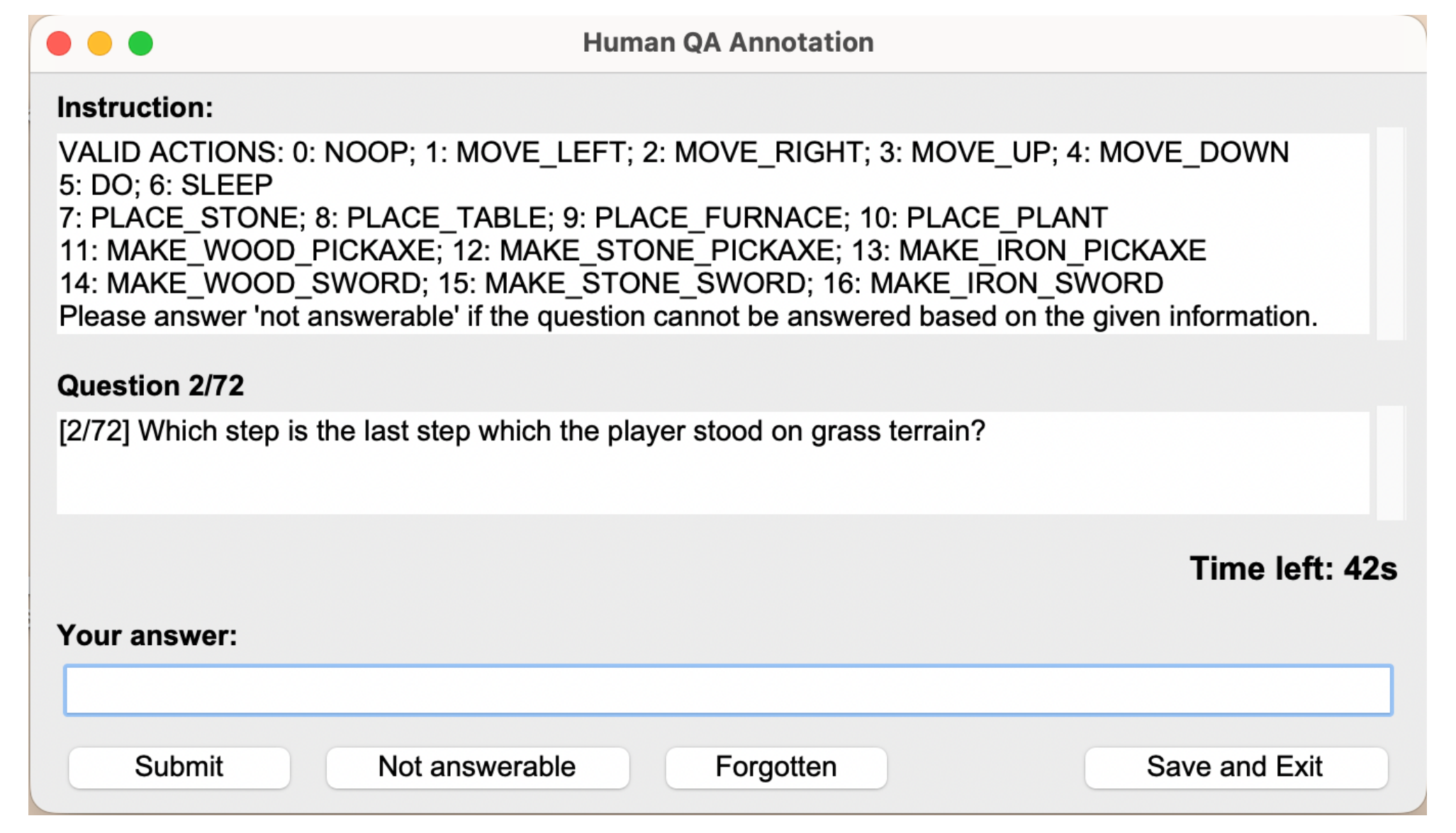}
  \caption{UI interface of the application for human evaluation question answering.}
  \label{fig:human_eval_ui}
\end{figure}

\paragraph{Interface and timing.}
Figure~\ref{fig:human_eval_ui} shows the QA interface. Each question has a 60-second time limit; timed-out questions are counted as incorrect. On average, annotators spend $\sim$40 minutes playing each text-game episode and $\sim$3 minutes for each visual-game episode; question answering takes $\sim$15 minutes (closed-book) and $\sim$30 minutes (open-book).

\begin{table}[t]
  \centering
  \scriptsize
  \setlength{\tabcolsep}{3pt}
  \resizebox{\columnwidth}{!}{%
    \begin{tabular}{lccc ccc}
      \toprule
      & \multicolumn{3}{c}{Open-book} & \multicolumn{3}{c}{Closed-book} \\
      \cmidrule(lr){2-4}\cmidrule(lr){5-7}
      Type & Acc(\%) & F1(\%) & Time(s) & Acc(\%) & F1(\%) & Time(s) \\
      \midrule
      Single-Hop  & 85.5 & 88.3 & 18.1 &  8.1 &  8.1 &  5.6 \\
      Multi-Hop   & 38.1 & 38.1 & 38.3 &  0.0 &  0.0 &  6.7 \\
      Induction   & 50.0 & 51.4 & 15.0 & 14.3 & 14.6 &  6.2 \\
      Spatial     & 34.7 & 38.4 & 20.1 &  5.7 &  6.0 &  6.5 \\
      Temporal    & 41.7 & 46.2 & 29.6 & 16.7 & 17.1 & 17.5 \\
      Logical     & 54.4 & 55.2 & 19.8 & 18.1 & 18.4 &  7.9 \\
      Adversarial & 90.2 &  -   &  9.8 & 46.3 &  -   &  4.7 \\
      \midrule
      Overall     & 59.2 & 56.5 & 19.9 & 15.8 & 10.8 &  7.6 \\
      \bottomrule
    \end{tabular}%
  }
  \caption{Human performance (visual game) averaged over annotators per question type. Acc/F1 are in \%.}
  \label{tab:human_per_type_visual}
\end{table}

\begin{table}[t]
  \centering
  \scriptsize
  \setlength{\tabcolsep}{2.5pt}
  \resizebox{\columnwidth}{!}{%
    \begin{tabular}{lccc ccc}
      \toprule
      & \multicolumn{3}{c}{Open-book} & \multicolumn{3}{c}{Closed-book} \\
      \cmidrule(lr){2-4}\cmidrule(lr){5-7}
      Game & Acc(\%) & F1(\%) & Time(s) & Acc(\%) & F1(\%) & Time(s) \\
      \midrule
      advent   & 55.7 & 53.2 & 21.8 & 12.3 & 12.0 &  7.9 \\
      awaken   & 65.7 & 66.0 & 16.7 & 27.5 & 24.6 &  5.5 \\
      balances & 67.9 & 65.5 & 21.1 & 28.8 & 24.1 & 10.3 \\
      dragon   & 65.6 & 65.2 & 21.0 & 17.3 & 13.2 &  7.4 \\
      gold     & 75.9 & 76.1 & 21.7 & 14.1 & 12.6 & 10.8 \\
      jewel    & 62.9 & 62.5 & 26.7 & 22.1 & 20.7 & 10.9 \\
      karn     & 60.2 & 58.6 & 27.0 & 17.4 & 13.3 & 19.0 \\
      ludicorp & 57.1 & 52.8 & 33.2 & 21.0 & 14.5 & 19.9 \\
      moonlit  & 65.3 & 60.7 & 24.4 & 28.2 & 24.0 & 11.6 \\
      pentari  & 69.0 & 66.4 & 24.3 & 23.3 & 17.7 &  9.5 \\
      reverb   & 66.6 & 64.8 & 21.4 & 30.1 & 24.8 & 12.2 \\
      sorcerer & 72.5 & 71.6 & 21.8 & 28.0 & 25.0 & 10.3 \\
      zork1    & 64.4 & 62.5 & 25.0 & 26.8 & 23.0 & 11.2 \\
      zork2    & 63.1 & 61.8 & 22.8 & 28.1 & 27.4 &  9.3 \\
      zork3    & 71.4 & 68.3 & 20.4 & 24.1 & 22.0 &  6.9 \\
      \midrule
      Overall  & 65.6 & 63.8 & 23.3 & 23.2 & 19.8 & 10.8 \\
      \bottomrule
    \end{tabular}%
  }
  \caption{Human performance (text games) averaged over two annotators per game. Acc/F1 are in \%.}
  \label{tab:human_per_game}
\end{table}

\paragraph{Metrics and aggregation.}
We report Accuracy, F1, and per-question time, aggregated by question type (Tables~\ref{tab:human_per_type} and \ref{tab:human_per_type_visual}) and by game for text games (Table~\ref{tab:human_per_game}).

\begin{table}[t]
  \centering
  \scriptsize
  \setlength{\tabcolsep}{4pt}
  \begin{tabular}{lcccc}
    \toprule
    Metric & Pearson $r$ & ICC & MAE & RMSE \\
    \midrule
    Open-book per-type Acc   & 0.90 & 0.71 & 0.09 & 0.11 \\
    Open-book per-type Time  & 0.68 & 0.40 & 5.69 & 6.66 \\
    Closed-book per-type Acc & 0.97 & 0.95 & 0.03 & 0.05 \\
    Closed-book per-type Time& 0.79 & 0.40 & 6.73 & 6.86 \\
    \bottomrule
  \end{tabular}
  \caption{Inter-annotator agreement for text games.}
  \label{tab:human_agreement}
\end{table}

\begin{table}[t]
  \centering
  \scriptsize
  \setlength{\tabcolsep}{4pt}
  \begin{tabular}{lcccc}
    \toprule
    Metric & Pearson $r$ & ICC & MAE & RMSE \\
    \midrule
    Open-book per-type Acc   & 0.85 & 0.78 & 0.22 & 0.30 \\
    Open-book per-type Time  & 0.62 & 0.45 & 5.82 & 6.54 \\
    Closed-book per-type Acc & 0.77 & 0.69 & 0.11 & 0.14 \\
    Closed-book per-type Time& 0.61 & 0.44 & 5.05 & 7.52 \\
    \bottomrule
  \end{tabular}
  \caption{Inter-annotator agreement for the visual game.}
  \label{tab:human_agreement_visual}
\end{table}

\paragraph{Inter-annotator agreement.}
We compute inter-annotator agreement over per-type aggregates using Pearson correlation ($r$), intraclass correlation (ICC), mean absolute error (MAE), and root mean squared error (RMSE) (Tables~\ref{tab:human_agreement} and \ref{tab:human_agreement_visual}).

\paragraph{Why closed-book is low while open-book is still imperfect.}
Closed-book accuracy is low because many questions require fine-grained episodic recall over long and partially observable trajectories, which exceeds ordinary human memory once external records are removed. Open-book access substantially improves performance by restoring access to evidence, but does not make the task trivial: annotators must still locate the relevant evidence in long logs and often need additional temporal, spatial, or aggregation reasoning to derive the final answer.

\paragraph{Additional interpretability analyses.}
Beyond aggregate accuracy and agreement, we further analyze why closed-book performance is low, why open-book performance remains imperfect, and how the 60-second limit affects outcomes. Concretely, we examine recurring open-book error types, a small time-limit sensitivity test, non-response / timeout / abstention behavior, and accuracy as a function of response time.

\subsection{Open-Book Error Analysis}

We focus on open-book errors because closed-book failures are much more dominated by non-response (e.g., ``cannot remember''). In the open-book setting, the trajectory serves as a memory aid rather than a fully structured document, and participants must still retrieve and integrate evidence under limited time. We observe five recurring error types.

\paragraph{Temporal anchor misalignment.}
Annotators sometimes fail on questions with explicit temporal anchors such as step $t$, before/after a step, or the first/last occurrence of an event. Typical mistakes include locating an adjacent segment, confusing before with after, or interpreting first/last as any occurrence.

\paragraph{Complex spatial trajectory reasoning.}
When a question requires multi-step movement tracing, nearest-target reasoning, shortest-path reasoning, or reachability checking, annotators may make one incorrect step in a long derivation or incorrectly decide that an answerable question is not answerable.

\paragraph{Counting and aggregation mistakes.}
For questions involving counts, distinct entities, most frequent events, or longest durations, annotators sometimes miss instances, double count, use inconsistent criteria, or abandon the attempt because the evidence is too dispersed.

\paragraph{Format mismatch.}
In some cases, annotators identify the right content but answer in a form that does not exactly match the required output format, such as giving a paraphrase instead of the exact observation span.

\paragraph{Misinterpretation of game dynamics.}
Some mistakes arise from misunderstanding game mechanics or conflating correlated cues with the target concept. For example, a participant may count all frames with nearby enemies instead of counting only actual attack events.

Representative examples are shown below.

\begin{quote}
\textbf{Temporal anchor misalignment (text, Logical)}\\
Q: After completing which step does the inventory contain `jewel' for the first time?\\
GT: 66\\
Human: 179
\end{quote}

\begin{quote}
\textbf{Complex spatial trajectory reasoning (visual, Spatial)}\\
Q: At step 100, starting from the player's current position, how should the player move to reach the nearest coal? Answer in `X steps left/right and Y steps up/down', or `not answerable' if never see the target, or 0 if player is on the target.\\
GT: 5 steps up, 4 steps left, 2 steps up and 1 step left\\
Human: (time-out)
\end{quote}

\begin{quote}
\textbf{Counting / aggregation mistake (visual, Induction)}\\
Q: How many distinct locations were visited in this game trajectory?\\
GT: 46\\
Human: not answerable
\end{quote}

\begin{quote}
\textbf{Format mismatch (text, Single-Hop)}\\
Q: At step 334, what observation did the player see?\\
GT: Maze. You are in a twisting maze of little passages, all alike.\\
Human: a maze of little twisting passages
\end{quote}

\begin{quote}
\textbf{Misinterpretation of game dynamics (visual, Logical)}\\
Q: How many times was the player attacked in the game?\\
GT: 7\\
Human: 14
\end{quote}

\begin{table}[t]
  \centering
  \scriptsize
  \setlength{\tabcolsep}{6pt}
  \begin{tabular}{lcc}
    \toprule
    Setting & Acc(\%) & F1(\%) \\
    \midrule
    Open-book, 60s  & 65.6 & 63.8 \\
    Open-book, 180s & 71.6 & 68.4 \\
    \midrule
    Closed-book, 60s  & 23.2 & 19.8 \\
    Closed-book, 180s & 25.1 & 20.4 \\
    \bottomrule
  \end{tabular}
  \caption{Time-limit sensitivity on a subset of 5 sampled text games. Extending the time budget from 60s to 180s yields a clear gain for open-book answering, but only a small gain for closed-book answering.}
  \label{tab:human_time_sensitivity}
\end{table}

\subsection{Time-Limit Sensitivity}

We impose a per-question time limit to prevent the evaluation from becoming unconstrained reading comprehension over lengthy logs. To test whether the 60-second limit is the main reason for low human performance, we rerun a subset of five sampled text games with a relaxed 180-second limit. The results are shown in Table~\ref{tab:human_time_sensitivity}.

The results show that extending the time budget yields only a small gain in the closed-book setting, but a larger gain in the open-book setting. This suggests that low closed-book scores are not mainly caused by the 60-second constraint: many closed-book failures arise because participants quickly decide that they cannot recall the needed evidence. By contrast, open-book answering benefits from extra time for locating and verifying evidence, especially for global aggregation and multi-step reasoning questions, but performance remains well below 100\%, indicating that evidence search is only part of the difficulty.

\begin{table}[t]
  \centering
  \scriptsize
  \setlength{\tabcolsep}{4pt}
  \resizebox{\columnwidth}{!}{%
  \begin{tabular}{lccccc}
    \toprule
    Dataset / Setting & Total Q & Non-response & Timeout & Answered-only Acc & False abstention \\
    \midrule
    Jericho open-book   & 1806 & 6.6\%  & 3.0\% & 66.8 & 2.7\% \\
    Jericho closed-book & 1806 & 51.1\% & 0.9\% & 48.0 & 9.0\% \\
    Crafter open-book   & 596  & 16.1\% & 2.0\% & 68.8 & 11.5\% \\
    Crafter closed-book & 596  & 72.1\% & 0.0\% & 55.4 & 19.0\% \\
    \bottomrule
  \end{tabular}%
  }
  \caption{Non-response, timeout, answered-only accuracy, and false-abstention statistics for human evaluation. Closed-book failures are dominated by rapid non-response rather than timeouts, while open-book access mainly helps by reducing skipping and restoring access to evidence.}
  \label{tab:human_nonresponse}
\end{table}

\subsection{Non-response, Timeout, and Abstention Analysis}

To characterize how humans fail under open-book and closed-book conditions, we separate attempt quality from non-response behavior. Specifically, we report: (i) \textit{non-response}, the fraction of questions with an empty answer or explicit ``cannot remember''; (ii) \textit{timeout}, the fraction of questions that exceed the time limit and yield no final answer; (iii) \textit{answered-only accuracy}, computed over questions with a non-empty answer; and (iv) \textit{false abstention}, the fraction of questions whose ground truth is answerable but the annotator responded ``not answerable''.

Table~\ref{tab:human_nonresponse} shows that closed-book failures are dominated by non-response rather than timeouts. For Jericho, closed-book non-response reaches 51.1\% with only 0.9\% timeouts; for Crafter, closed-book non-response reaches 72.1\% with 0.0\% timeouts. This indicates that participants often skip quickly when they cannot recall the relevant evidence, rather than spending the full time budget. Open-book access sharply reduces non-response and improves answered-only accuracy, especially in Jericho, showing that access to the trajectory mainly helps by restoring evidence and enabling verification. We also observe higher false-abstention rates in closed-book settings, suggesting that participants often forget the existence of relevant events or objects and therefore judge answerable questions as not answerable.

\begin{table}[t]
  \centering
  \scriptsize
  \setlength{\tabcolsep}{4pt}
  \resizebox{\columnwidth}{!}{%
  \begin{tabular}{lccc}
    \toprule
    Time bin & All-questions Acc & Answered-only Acc & Unanswered rate \\
    \midrule
    \multicolumn{4}{l}{\textbf{Jericho open-book}} \\
    0--15s  & 76.3 & 80.9 & 5.6\% \\
    15--30s & 62.8 & 64.3 & 2.4\% \\
    30--45s & 54.0 & 55.3 & 2.4\% \\
    45--60s & 40.0 & 42.8 & 6.5\% \\
    \midrule
    \multicolumn{4}{l}{\textbf{Jericho closed-book}} \\
    0--15s  & 24.0 & 56.0 & 57.1\% \\
    15--30s & 24.0 & 32.3 & 25.9\% \\
    30--45s & 21.7 & 30.0 & 27.5\% \\
    45--60s & 10.7 & 12.5 & 14.3\% \\
    \bottomrule
  \end{tabular}%
  }
  \caption{Jericho human accuracy by response-time bucket. In open-book mode, longer times correlate with lower accuracy, consistent with harder items taking longer. In closed-book mode, low overall accuracy is driven largely by rapid abstention rather than fast incorrect attempts.}
  \label{tab:human_time_bucket_jericho}
\end{table}

\begin{table}[t]
  \centering
  \scriptsize
  \setlength{\tabcolsep}{4pt}
  \resizebox{\columnwidth}{!}{%
  \begin{tabular}{lccc}
    \toprule
    Time bin & All-questions Acc & Answered-only Acc & Unanswered rate \\
    \midrule
    \multicolumn{4}{l}{\textbf{Crafter open-book}} \\
    0--15s  & 49.3 & 68.8 & 28.4\% \\
    15--30s & 57.7 & 63.8 & 9.6\% \\
    30--45s & 58.8 & 62.5 & 5.9\% \\
    45--60s & 18.2 & 22.2 & 18.2\% \\
    \midrule
    \multicolumn{4}{l}{\textbf{Crafter closed-book}} \\
    0--15s  & 17.0 & 51.1 & 66.7\% \\
    15--30s & 16.7 & 50.0 & 66.7\% \\
    30--45s & 0.0  & 0.0  & 33.3\% \\
    45--60s & 50.0 & 50.0 & 0.0\% \\
    \bottomrule
  \end{tabular}%
  }
  \caption{Crafter human accuracy by response-time bucket. Even in open-book mode, short-bin skipping is more common than in Jericho, likely reflecting the higher verification burden in a visually grounded environment. In closed-book mode, most failures are best interpreted as abstention rather than timeout.}
  \label{tab:human_time_bucket_crafter}
\end{table}

\subsection{Accuracy by Response-Time Bucket}

We further bucket questions by response time into 0--15s, 15--30s, 30--45s, and 45--60s, and report all-questions accuracy, answered-only accuracy, and unanswered rate for each bin (Tables~\ref{tab:human_time_bucket_jericho} and \ref{tab:human_time_bucket_crafter}).

For Jericho open-book, unanswered rates remain low, and longer response times correlate with lower accuracy, consistent with the standard pattern that more difficult items take longer and remain more error-prone. For Jericho closed-book, overall accuracy is low largely because many responses in the shortest bins are immediate abstentions, while answered-only accuracy is substantially higher whenever participants do attempt an answer. Crafter shows a similar but stronger pattern: even in open-book mode, short-bin skipping is more common than in Jericho, likely due to the higher verification burden in a visually grounded environment. In closed-book Crafter, most failures are best characterized as abstention rather than timeout.

%% file: 9_appendix_crossbench.tex
\section{Cross-Benchmark Comparison Details}
\label{app:crossbench}

\begin{table}[t]
  \centering
  \scriptsize
  \setlength{\tabcolsep}{4pt}
  \begin{tabular}{l cc cc cc}
    \toprule
    & \multicolumn{2}{c}{Single-Hop} & \multicolumn{2}{c}{Multi-Hop} & \multicolumn{2}{c}{Temporal} \\
    \cmidrule(lr){2-3}\cmidrule(lr){4-5}\cmidrule(lr){6-7}
    Method & Ours & LoCoMo & Ours & LoCoMo & Ours & LoCoMo \\
    \midrule
    Mem0    & +1.60 & +4.42 & -27.70 & +8.55  & -13.90 & +34.89 \\
    LangMem & -1.70 & +1.21 & -14.80 & +5.95  &  -7.10 & +16.71 \\
    A-MEM   & +17.90 & -7.28 &  -4.60 & -7.95 & -11.60 & +31.81 \\
    \bottomrule
  \end{tabular}
  \caption{\textbf{Type-wise gain comparison on three shared reasoning types.}
  Each entry is an \emph{absolute gain in points} relative to an in-context prompting baseline within the same benchmark.
  A leading ``+'' means the method improves the score, and ``-'' means it decreases.}
  \label{tab:ours_vs_locomo_gains_three_types}
\end{table}

In \cref{sec:controlled_exposure_eval}, we show that the fixed-QA setting can change the relative advantages of memory methods. This motivates a cross-benchmark comparison: to what extent do our findings align with existing fixed memory benchmarks?

We use Qwen-family models and compare our results with benchmarks \textbf{LongMemEval} and \textbf{LoCoMo}. Since these benchmarks differ in task form and metrics, we emphasize \emph{relative} improvements over a within-benchmark baseline rather than absolute scores.
Formally, we compare ability-level gain vectors
$\Delta \mathbf{a}(m) = \mathbf{a}(m) - \mathbf{a}(m_0)$,
where $\mathbf{a}(m)$ is the vector of type-wise scores and $m_0$ is the in-context baseline.

\paragraph{Agreement in overall ordering.}
On LongMemEval, a Qwen3-based unified comparison shows A-MEM outperforming FullText, while LangMem and Mem0 trail behind, suggesting that more structured/agentic memory can be advantageous for assistant-style long-term memory.
This qualitative ordering is consistent with our Qwen3-32B text-game results, where A-MEM is also the strongest overall method among the shared set.

\paragraph{Mismatch in \emph{where} gains come from.}
Table~\ref{tab:ours_vs_locomo_gains_three_types} reveals a clear mismatch in \emph{where} memory helps. On LoCoMo, temporal questions largely reduce to reconstructing cross-session timelines from retrievable dialogue facts, so external memory and retrieval tend to be directly beneficial.
In our benchmark, ``temporal'' questions more often query step-level ordering and state evolution grounded in an interactive environment, where evidence is partially observed across many turns; compressive memory and imperfect retrieval can therefore blur fine-grained ordering, hurting temporal accuracy relative to in-context reasoning. But we also notice in Table \ref{tab:main-experiments} that A-MEM can improve performance for temporal questions for strong models like GPT-5.1.
\FloatBarrier

%% file: 9_appendix_horizoncontrol.tex
\section{Full Results of Query Horizon Control Setting}\label{app:queryhorizon}

\begin{table*}
    \centering
    \scriptsize
    \setlength{\tabcolsep}{2pt}
    \resizebox{\textwidth}{!}{%
    \begin{tabular}{ll*{9}{c}}
        \toprule
        \multirow{2}{*}{Model} & \multirow{2}{*}{Method} &
        \multicolumn{1}{c}{Single-Hop} &
        \multicolumn{1}{c}{Multi-Hop} &
        \multicolumn{1}{c}{Induction} &
        \multicolumn{1}{c}{Spatial} &
        \multicolumn{1}{c}{Temporal} &
        \multicolumn{1}{c}{Logical} &
        \multicolumn{1}{c}{Adversarial} &
        \multicolumn{2}{c}{Overall} \\
        \cmidrule(lr){3-3}
        \cmidrule(lr){4-4}
        \cmidrule(lr){5-5}
        \cmidrule(lr){6-6}
        \cmidrule(lr){7-7}
        \cmidrule(lr){8-8}
        \cmidrule(lr){9-9}
        \cmidrule(lr){10-11}
        & &
        ACC &
        ACC &
        ACC &
        ACC &
        ACC &
        ACC &
        ACC &
        ACC & F1 \\
        \midrule
        \multicolumn{11}{c}{\textbf{Text-only Games}} \\
        \midrule
        \multirow{4}{*}{GPT-5.1}
            & In-context   & \textbf{71.8} & 34.7 & \textbf{40.5} & 46.0 & 51.7 & 43.3 & 43.0 & 49.3 & \underline{48.3}  \\
            & Mem0         & 51.9 & 8.6 & 16.7 & 14.6 & 53.5 & 25.2 & 84.5 & 37.7 & 34.7 \\
            & LangMem      & 44.3 & 34.5 & 22.5 & \underline{47.5} & \textbf{72.3} & 34.5 & 81.4 & 48.6 & 45.6 \\
            & A-MEM        & 48.5 & 33.9 & 27.2 & 47.0 & 59.0 & \underline{62.0} & 74.4 & 49.6 & 46.0  \\
        \midrule
        \multirow{4}{*}{Qwen2.5-32B-Instruct}
            & In-context   & 52.3 & 23.1 & \underline{38.4} & 34.3 & 52.5 & 41.9 & 50.0 & 41.0 & 40.1 \\
            & Mem0         & 54.7 & 8.0 & 14.3 & 8.6 & 47.1 & 22.2 & \textbf{89.9} & 36.4 & 32.5 \\
            & LangMem      & 48.1 & \underline{37.6} & 31.7 & \textbf{50.3} & 55.6 & 28.2 & 84.9 & 49.9 & 46.5 \\
            & A-MEM        & \underline{67.3} & 30.7 & 26.2 & 45.6 & 58.0 & \textbf{65.4} & \underline{88.7} & \textbf{54.7} & \textbf{51.0} \\
        \midrule
        \multirow{4}{*}{Qwen3-32B}
            & In-context   & 57.8 & \textbf{46.7} & 33.6 & 47.1 & 48.5 & 41.3 & 46.2 & 47.7 & 46.9 \\
            & Mem0         & 54.3 & 1.9 & 24.2 & 5.2 & \underline{65.5} & 24.3 & 84.9 & 41.3 & 33.7 \\
            & LangMem      & 45.5 & 17.0 & 25.6 & 44.3 & 48.0 & 34.8 & 71.0 & 41.4 & 36.0 \\
            & A-MEM        & 61.6 & 35.7 & 25.1 & 42.5 & 48.7 & 59.6 & 75.6 & \underline{50.8} & 46.8 \\
        \midrule
        \multicolumn{11}{c}{\textbf{Visual Games}} \\
        \midrule
        \multirow{2}{*}{GPT-5.1}
            & In-context   & 59.5 & 38.5 & 39.0 & 26.6 & \underline{56.7} & 36.1 & \textbf{64.6} & \textbf{48.2} & \textbf{46.0} \\
            & A-MEM        & 54.8 & \textbf{46.0} & 34.8 & \textbf{32.6} & \textbf{60.7} & 31.0 & 55.8 & 46.8 & 42.3 \\
        \cmidrule{1-11}
        \multirow{2}{*}{Qwen3VL-32B-Instruct}
            & In-context   & \underline{59.6} & \underline{41.8} & \textbf{39.6} & 24.9 & 36.7 & 48.2 & 62.4 & 47.1 & 42.8 \\
            & A-MEM        & \textbf{62.8} & 38.7 & \underline{39.3} & \underline{27.0} & 32.9 & \textbf{52.3} & \underline{62.8} & \underline{48.0} & \underline{45.0} \\
        \cmidrule{1-11}
        \multirow{2}{*}{InternVL3.5-38B}
            & In-context   & 34.1 & 10.7 & 17.4 & 12.3 & 34.1 & 44.7 & 48.5 & 30.9 & 26.4 \\
            & A-MEM        & 38.4 & 7.3 & 18.1 & 14.4 & 36.9 & \underline{48.8} & 55.4 & 33.9 & 28.7 \\
        \bottomrule
    \end{tabular}%
    }
    \caption{Query-horizon control results with queries restricted to steps 1--50. Within each modality block, the best and second-best numbers in each column are \textbf{bolded} and \underline{underlined}, respectively.}
    \label{tab:main-experiments-50rounds}
\end{table*}

We provide the full results of query horizon control setting in Table~\ref{tab:main-experiments-50rounds}. 

\FloatBarrier

%% file: 9_appendix_casestudy.tex
\section{Case Study}
\label{app:case_studies}

This appendix presents two complementary groups of case studies for memory agents. We present \textbf{five common mistakes}—(1) evidence localization errors, (2) long-horizon aggregation failures, (3) visual grounding mistakes, (4) reasoning confusion, and (5) answer--explanation inconsistency. Additionally, we provide \textbf{two challenging question analyses} that are hard-to-solve for memory agents so far: (6) global object counting under partial observability and (7) target-reaching calculation with obstacles and detours.

\CaseTitle{Case 1: Evidence localization error.}
\noindent\Qline{What action is taken 2 steps after the first step where the reason mentions `out'?}\\
\GTline{north}\\
\Predline{Not answerable.}\\
\Expline{The snippets only provide selected reasoning texts, observations, and results for certain steps, but not a complete mapping from every step number to its action or full reasoning text. To answer the question, we would need to (1) identify the earliest step whose reasoning includes the word `out', and then (2) know the exact action taken two steps after that step.}\\
\Ctxblock{%
t=79 \;|\; action=east \;|\; reason=Try unexplored maze direction to find a way out \;|\;
observation=\textbackslash nMaze\textbackslash n\textbackslash nA threatening little dwarf comes out of the shadows! \;|\;
location=(Alike\_Maze\_3) \;|\; reward=32.0 \;|\; done=false\\
t=80 \;|\; action=west \;|\; reason=Try a new maze direction; avoid repeating recent moves \;|\;
observation=\textbackslash nMaze\textbackslash n\textbackslash nThe dwarf stalks after you...\textbackslash n\textbackslash nThe dwarf throws a nasty little knife at you, but misses! \;|\;
location=(Alike\_Maze\_2) \;|\; reward=32.0 \;|\; done=false\\
t=81 \;|\; action=north \;|\; reason=Keep exploring maze; try to find exit or landmark \;|\;
observation=\textbackslash nMaze\textbackslash n\textbackslash nA threatening little dwarf hides in the shadows.\textbackslash n\textbackslash nThe dwarf throws a nasty little knife at you, but misses! \;|\;
location=(Alike\_Maze\_1) \;|\; reward=32.0 \;|\; done=false
}

\noindent\textbf{Error analysis.} This is an evidence retrieval failure over step-indexed reasoning. The correct anchor is the first step whose \emph{reason} contains the keyword `out' (t=79), and the answer requires a deterministic offset over consecutive steps (t=81 $\rightarrow$ \texttt{north}). The agent cannot reliably retrieve the earliest keyword-bearing reason with its timestep and the subsequent actions. Such multi-hop queries are difficult: missing a single anchor step makes the downstream step lookup appear unsupported, resulting in a false ``Not answerable'' decision.

\CaseTitle{Case 2: Answer--explanation inconsistency.}
\noindent\Qline{What was the longest unbroken sequence of \texttt{MOVE\_LEFT} actions?}\\
\GTline{4}\\
\Predline{3}\\
\Expline{The longest consecutive sequence of MOVE\_LEFT actions is from t=97--t=100 (97,98,99,100) which is 4? Wait; checking: 97,98,99,100 are four in a row.}\\

\Ctxblock{%
t=97 \;|\; action=MOVE\_LEFT \;|\; frame=crafter\_t000097.png \;|\; (health,food,water,energy)=(9,6,5,6)\\
reason=Needs are stable and we want to reach the stone wall on the far left; move left while facing left to start closing the distance toward the stone for mining with our wood pickaxe while staying near the lake for later drinking.\\
t=98 \;|\; action=MOVE\_LEFT \;|\; frame=crafter\_t000098.png \;|\; (health,food,water,energy)=(9,6,5,6)\\
reason=Needs are stable and we are close to the stone wall; move left while facing left to stand directly adjacent to the stone so we can start mining it with DO using our wood pickaxe, while still remaining near the lake for later drinking.\\
t=99 \;|\; action=MOVE\_LEFT \;|\; frame=crafter\_t000099.png \;|\; (health,food,water,energy)=(9,6,5,6)\\
reason=Needs are fine and the stone wall is just to our left; step left while facing left to become directly adjacent so we can start mining stone with DO next, staying close to the lake for future drinking.\\
t=100 \;|\; action=MOVE\_LEFT \;|\; frame=crafter\_t000100.png \;|\; (health,food,water,energy)=(9,6,5,6)\\
reason=Status is safe and the stone wall is immediately to our left; move left while facing left to stand adjacent to the stone so we can start mining it with DO using our wood pickaxe, still remaining close to the lake for later drinking.
}

\noindent\textbf{Error analysis.} The explanation itself identifies a 4-step run, but the final answer reports 3, reflecting a self-contradiction between the produced rationale and the emitted prediction. For integer-valued questions, such slips (off-by-one or inconsistent finalization) cause complete loss under exact-match scoring even when the intermediate reasoning is correct.

\CaseTitle{Case 3: Long-horizon aggregation failure.}
\noindent\Qline{How many unique locations have been visited?}\\
\GTline{25}\\
\Predline{18}\\
\Expline{We found these distinct location names: Narrow Tunnel, Foot Bridge, Great Cavern, Shallow Ford, Dark Tunnel, Path Near Stream, Ice Room, Cool Room, Inside the Barrow, Bank Entrance, Fresco Room, West Viewing, East Viewing, Lava Tube, Formal Garden, Dragon Room, Stone Bridge, Gazebo, Carousel Room, North End of Garden. After deduplication (e.g., Ice vs Ice Room; Fresco vs Fresco Room) we get 18.}\\
\Ctxblock{All locations that appear in the recorded map: 1.~Inside Barrow; 2.~Narrow Tunnel; 3.~Foot Bridge; 4.~Shallow Ford; 5.~Dark Tunnel; 6.~Path Near Stream; 7.~Formal Garden; 8.~North End Garden; 9.~Gazebo; 10.~GreCavern; 11.~Topiary; 12.~Carousel; 13.~Cool; 14.~Ice; 15.~Lava Tube; 16.~Cobwebby Corridor; 17.~Stone Bridge; 18.~Drag; 19.~End Ledge; 20.~Fresco; 21.~Bank Entrance; 22.~West Teller's; 23.~West Viewing; 24.~East Teller's; 25.~East Viewing.}
\noindent\textbf{Error analysis.} This combines (i) long-horizon set aggregation—omissions accumulate silently over many steps; and (ii) string-level dedup ambiguity—surface-form heuristics can over-merge distinct nodes or miss true duplicates. When the number of visited locations grows, both omission and mis-dedup become more likely, producing systematic undercounts even when the agent attempts an explicit list.

\begin{figure}[h]
  \centering
  \includegraphics[width=0.50\linewidth]{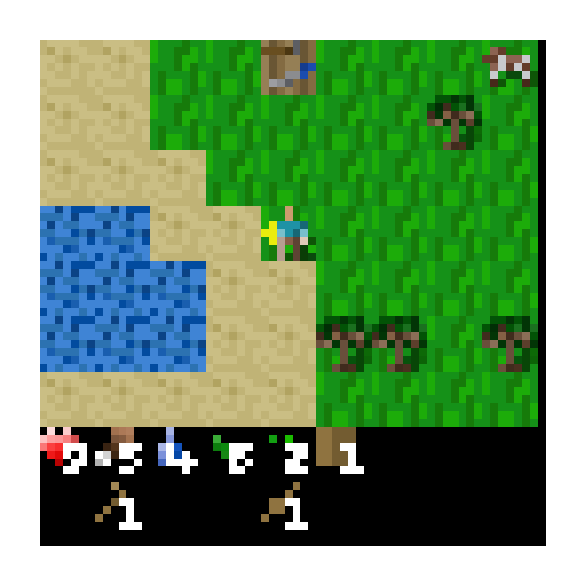}
  \caption{Frame for step 113.}
  \label{fig:step113}
\end{figure}

\CaseTitle{Case 4: Visual grounding mistake.}
\noindent\Qline{What terrain was under the player at step 113?}\\
\GTline{grass}\\
\Predline{Sand}\\
\Expline{Reasoning for t=113 says the player is standing on sand directly beside the lake.}\\
\Ctxline{Figure~\ref{fig:step113}}\\
\\
\noindent\textbf{Error analysis.} The model confuses \emph{nearby salient terrain} with the \emph{tile under the player}. Visually, sand may be adjacent and prominent, but the question requires a fine-grained under-agent grounding decision; the prediction follows the plausible nearby cue rather than the precise queried support.

\FloatBarrier

\begin{figure}[h]
  \centering
  \includegraphics[width=0.65\linewidth]{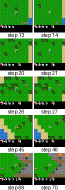}
  \caption{Steps that agent successfully collected wood.}
  \label{fig:stepcollecttree}
\end{figure}

\CaseTitle{Case 5: Reasoning Confusion.}
\noindent\Qline{How many times did the player successfully collect wood?}\\
\GTline{5}\\
\Predline{20}\\
\Expline{There are 20 DO actions explicitly described as chopping trees to collect or gain more wood.}\\
\Ctxline{Figure~\ref{fig:stepcollecttree}}\\
\\
\noindent\textbf{Error analysis.} The model counts \emph{attempts} (described chopping/DO actions) instead of \emph{verified successful outcomes}. In interactive settings, attempting to collect a resource does not guarantee success; success is better indicated by state-change evidence (e.g., wood count increases) or explicit success events. Treating attempts as successes inflates the count whenever failures occur.

\begin{figure}[h]
  \centering
  \includegraphics[width=0.75\linewidth]{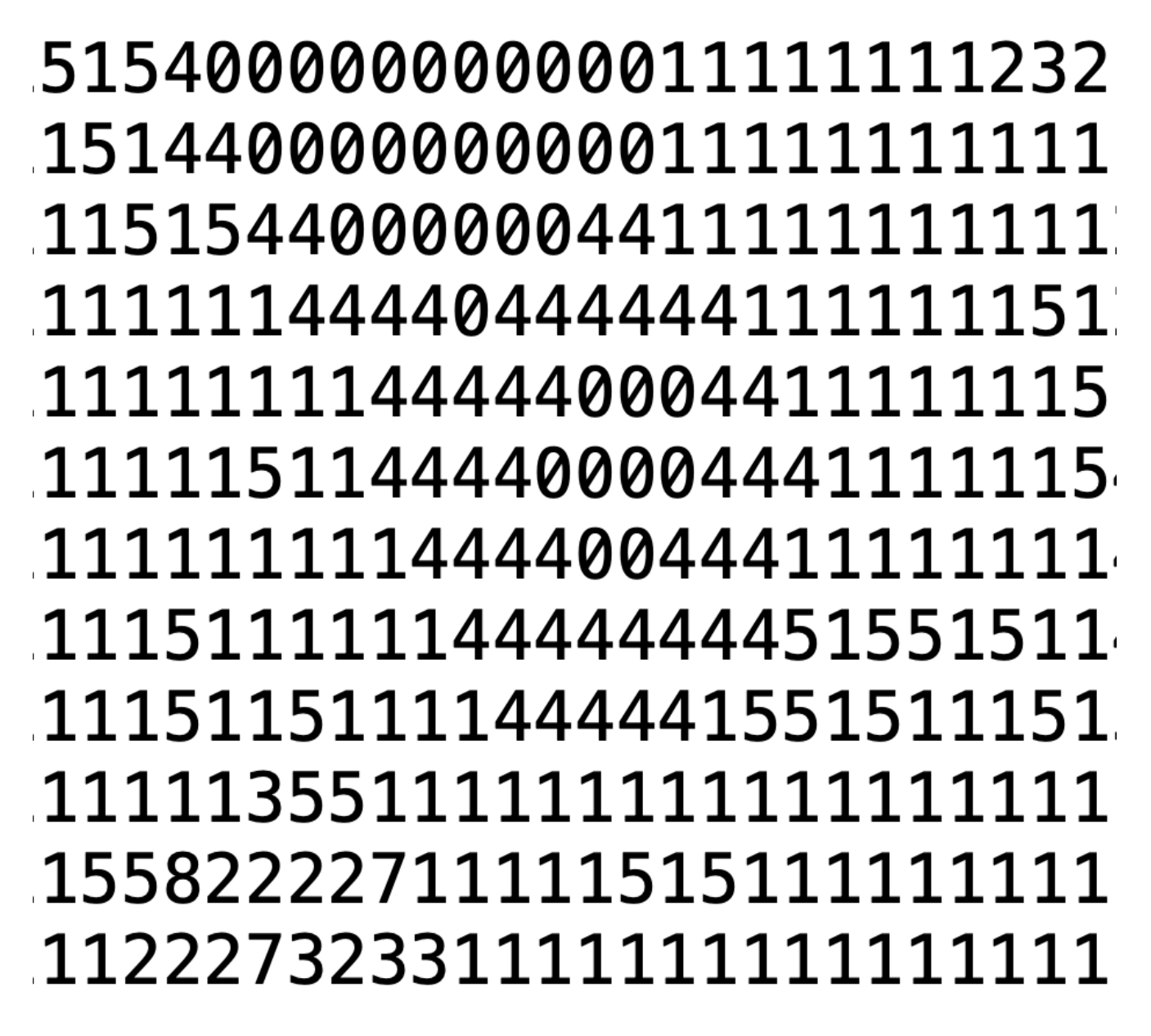}
  \caption{Part of the global map in backend data, full map is 64x64. Each number represents a type of tile.}
  \label{fig:globalmap}
\end{figure}

\begin{figure}[h]
  \centering
  \includegraphics[width=0.95\linewidth]{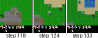}
  \caption{Figure for frames near the lake.}
  \label{fig:frames133}
\end{figure}

\CaseTitle{Case 6: Challenging Question 1.}
\noindent\Qline{What is the total number of unique trees the player encountered?}\\
\GTline{18}\\
\Predline{Not answerable.}\\
\Expline{The 64$\times$64 view changes as the player moves, so the same tree can leave and re-enter the frame; without exact map coordinates we cannot distinguish unique trees.}\\
\Ctxline{Figure \ref{fig:globalmap}}\\
\\
\textbf{Difficulty analysis.} This question is challenging because the agent observes only a partial map at each step. As a result, counting ``unique trees'' cannot be solved by summing per-frame tree counts; the same tree may repeatedly enter/leave the view across time. Answering correctly requires strong spatial reasoning to (i) integrate observations over time into a coherent global map, and (ii) perform object identity tracking across frames to deduplicate revisits. Although the agent refuses to answer, the question is in fact answerable. In our ground-truth pipeline, we first extract a global map for each game instance \textbf{(0=water, 1=grass, 2 = stone, 3 = coal, 4 = sand, 5=tree, etc.)}. Using the agent's real-time position, we compute the union of all map regions that were ever observable along the trajectory, i.e., the exact explored/seen area. Meanwhile, we update the global map according to the agent's actions (e.g., chopping a tree), so the global map reflects the environment dynamics rather than a static snapshot. With the reconstructed seen-area plus these action-conditioned updates, the total number of unique encountered trees can be calculated.
\clearpage
\CaseTitle{Case 7: Challenging Question 2 .}
\noindent\Qline{At step 118, from the player's current location, what movement is required to reach the nearest lake? Respond with `X steps left/right and Y steps up/down', `not answerable' if the target is never visible, or 0 if the player is already at the target.}\\
\GTline{1 step right, 6 steps up and 3 steps right}\\
\Predline{4 steps right and 7 steps up}\\
\Expline{From step 118 the nearest visible lake is the one reached earlier near steps 131--133, offset about 4 tiles right and 7 up.}\\
\Ctxline{Figure \ref{fig:frames133}}\\
\\
\textbf{Difficulty analysis.} The agent successfully identifies a nearest lake that is not in the current frame, implying it can relate the current position to a previously observed landmark across time, which is already a non-trivial spatial reasoning capability. The remaining difficulty stems from two factors. First, the vertical distance is miscounted by one step (the correct upward movement is 6, not 7), showing how sensitive such questions are to small coordinate errors. Second, the direct ``up'' displacement is not necessarily traversable: \textbf{the tile immediately above the agent is stone and cannot be crossed}, so the valid shortest route must detour around the obstacle. The ground-truth computation follows the navigation algorithm in Appendix~\ref{app:gt_d_nav_to_target}, which explicitly accounts for impassable terrain when deriving the minimal path.

%% file: 9_appendix_jericho.tex
\section{Text-Game Environment (Jericho)}
\label{app:jericho}

\subsection{Jericho environment}
We evaluate long-horizon memory in \emph{interactive fiction} (IF), where an agent interacts with a world purely through natural-language commands (e.g., \texttt{take lantern}, \texttt{go north}). We use \textbf{Jericho}~\cite{jericho}, a lightweight Python interface that connects learning agents to classic Z-Machine story files (e.g., \texttt{.z3/.z5}) through a Gym-like API. At each step, the agent emits a text action; the environment returns a textual observation, a scalar reward/score signal (game-dependent), a termination flag, and auxiliary metadata. Jericho also provides utilities commonly used by text-game agents, such as querying valid actions and accessing structured state information exposed by the underlying interpreter (when available). This setup yields long trajectories with sparse rewards and a large, compositional action space, making it a suitable testbed for evaluating memory-dependent reasoning across extended interaction histories.

\begin{table*}[t]
  \centering
  \scriptsize
  \setlength{\tabcolsep}{4pt}
  \resizebox{\textwidth}{!}{%
  \begin{tabular}{llp{4.2cm}p{8.4cm}}
    \toprule
    \textbf{ROM} & \textbf{Title} & \textbf{Genre / Setting} & \textbf{One-sentence synopsis} \\
    \midrule
    advent.z5    & Colossal Cave Adventure & Classic exploration / cave-crawl & Explore a vast cave system, solve inventory puzzles, and collect treasures for score. \\
    awaken.z5    & The Awakening & Mystery / survival & You awaken amid storm and mud with fragmented memory and must recover your bearings and escape danger. \\
    balances.z5  & Balances & Fantasy puzzle / spellcraft & A short, spell-focused adventure: learn and cast scroll-based spells to progress and collect objectives. \\
    dragon.z5    & Dragon! & Micro-IF / humorous vignette & A very short scenario centered on playing a dragon pursuing a single-minded goal. \\
    gold.z5      & Goldilocks is a FOX! & Fairy-tale remix / puzzle adventure & A puzzle-driven mashup of fairy-tale motifs (e.g., bears, wolves, magic-tale references). \\
    jewel.z5     & The Jewel of Knowledge & Classic dungeon crawl & Brave a lethal multi-layer dungeon to obtain the fabled Jewel of Knowledge guarded by powerful foes. \\
    ludicorp.z5  & The Ludicorp Mystery & Modern mystery / office building & Investigate a game company’s office to uncover why a long-awaited release has gone missing. \\
    moonlit.z5   & The Moonlit Tower & Symbolic fantasy / atmospheric & Navigate a surreal tower of symbols (masks, seasons, symmetry) to uncover meaning and resolution. \\
    pentari.z5   & Pentari & Fantasy / military prequel & As a company commander on leave, you navigate a compact fantasy town and events before a looming mission. \\
    reverb.z5    & Reverberations & Thriller / comedy-adventure & A pizza-delivery surfer gets entangled in an escape plot, murder attempts, and a conspiracy with romantic side-threads. \\
    sorcerer.z3  & Sorcerer & Fantasy / Infocom spell adventure & Track the vanished guildmaster Belboz via magical clues and avert a looming threat to the Circle of Enchanters. \\
    zork1.z5     & Zork I: The Great Underground Empire & Classic treasure-hunt dungeon & Explore the Great Underground Empire, solve puzzles, and amass treasures for points. \\
    zork2.z5     & Zork II: The Wizard of Frobozz & Classic dungeon + antagonist & Continue deeper in Zork while contending with the Wizard’s interference and advanced puzzles. \\
    zork3.z5     & Zork III: The Dungeon Master & Classic dungeon / endgame trial & Push into the deepest reaches of Zork for a culminating trial involving the enigmatic Dungeon Master. \\
    \bottomrule
  \end{tabular}%
  }
  \vspace{2pt}
  \footnotesize\emph{Note:} Synopses are summarized from publicly available game catalog blurbs and archival entries (e.g., IFDB/IFWiki/IF Archive).
  \caption{Jericho text-game suite used in this paper (ROM filenames and brief synopses).}
  \label{tab:jericho_games}
\end{table*}

\subsection{Game suite}
Table~\ref{tab:jericho_games} lists the Jericho games used in our text-game benchmark. They cover diverse genres (classic cave-crawls, office mysteries, fantasy quests, short-form vignettes) and vary widely in exploration difficulty, puzzle dependency, and required world knowledge.

%% file: 9_appendix_crafter.tex
\section{Visual Game Environment (Crafter)}
\label{app:crafter}

\subsection{Crafter environment}
We also evaluate memory in a \emph{visual} open-world survival environment, \textbf{Crafter}~\cite{hafner2021crafter}, which is designed as a compact Minecraft-inspired benchmark with procedurally generated 2D worlds and a technology tree (resources $\rightarrow$ tools $\rightarrow$ advanced resources). Crafter episodes are long-horizon and partially observable: at each step, the agent receives a \textbf{$64\times 64$ RGB} top-down crop centered on the player and must balance survival (food, water, energy, health) with exploration and crafting progress~\cite{hafner2021crafter}. Agents are evaluated by whether they can unlock a diverse set of semantically meaningful achievements within an episode, which probes generalization, long-term reasoning, and deep exploration~\cite{spring2023}.

\begin{figure}[h]
  \centering
  \includegraphics[width=\linewidth]{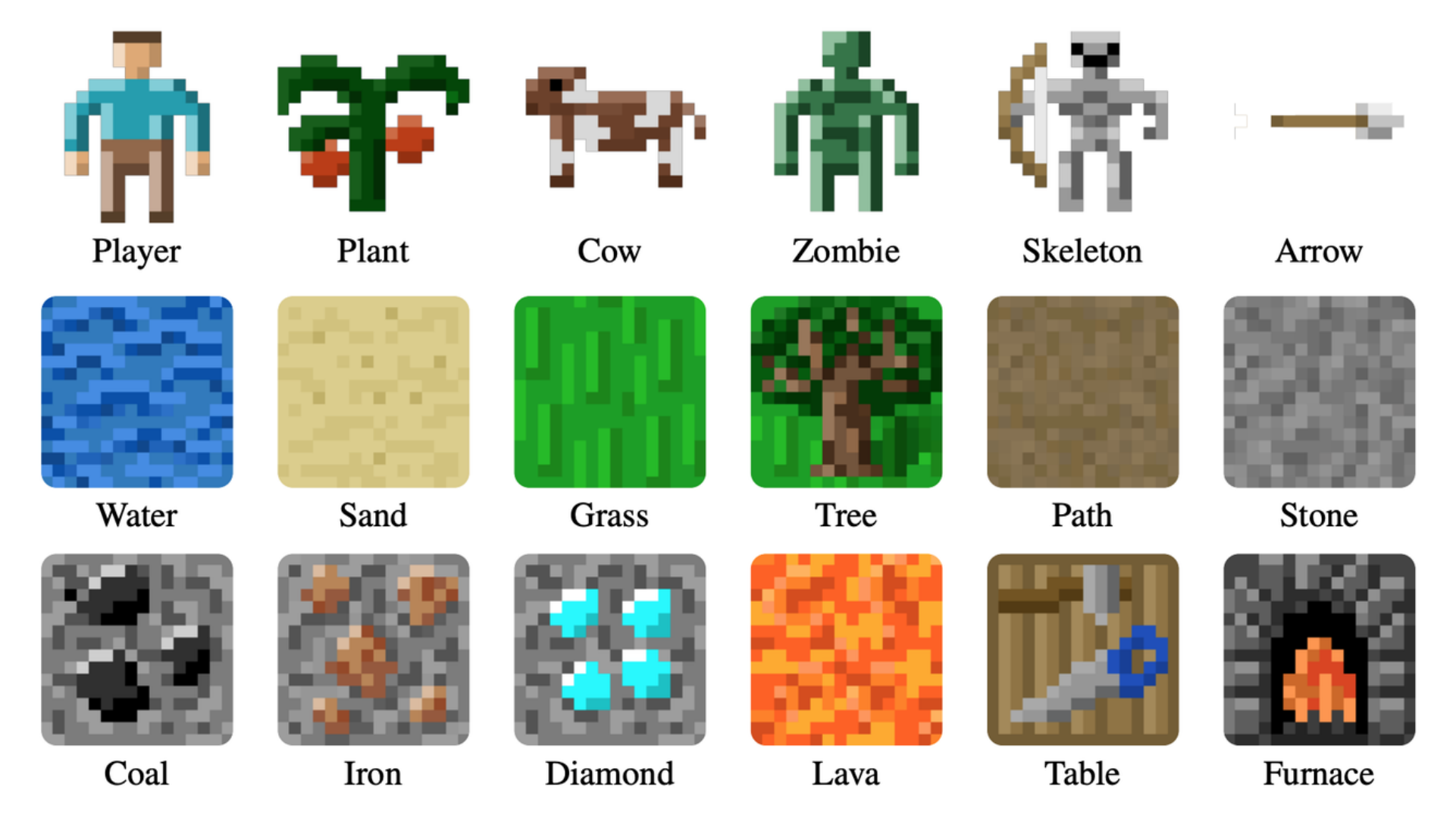}
  \caption{Crafter icon legend used by our visual agents. We attach this legend to each step to disambiguate terrain, resources, stations, and creatures from map.}
  \label{fig:crafter_icons}
\end{figure}

In our setup, we provide a textual HUD summary each step (derived from the environment state) and attach a fixed icon legend to help resolve object identities from pixels. Figure~\ref{fig:crafter_icons} shows the icon legend used throughout our experiments.

\subsection{Observation example}
\begin{figure}[h]
  \centering
  \includegraphics[width=0.52\linewidth]{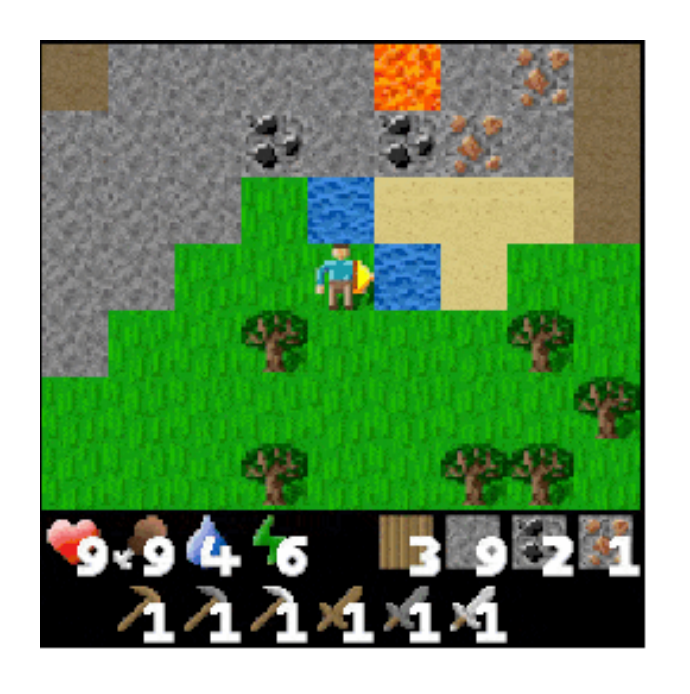}
  \caption{Example Crafter observation (agent input). The agent receives a $64\times64$ RGB image containing a local top-down crop around the player (top) and an inventory/status HUD (bottom).}
  \label{fig:crafter_obs_example}
\end{figure}

Figure~\ref{fig:crafter_obs_example} shows a single Crafter observation as seen by the agent.
The \emph{upper} region is a local top-down view centered on the player, while the \emph{bottom} HUD
encodes the player’s internal state (e.g., health/food/water/rest) and inventory counts (materials and crafted tools).
In this example, the player stands at the boundary between grass and water, with nearby trees and stone terrain containing coal and lava. Throughout our experiments, we additionally provide a textual HUD summary each step (for exact numbers) and attach the icon legend in Fig.~\ref{fig:crafter_icons} to disambiguate object identities.

\subsection{Crafter instructions}
\begin{tcolorbox}[
  title=\textbf{Crafter Instructions},
  colback=black!2,
  colframe=black!25,
  boxrule=0.5pt,
  arc=2pt,
  left=6pt,right=6pt,top=6pt,bottom=6pt
]
\footnotesize
\textbf{WORLD VIEW \& HUD.}
You aim to survive longer and maximize unlocked achievements.
You observe a $64\times 64$ RGB top-down crop centered on the player. A HUD exists in the image; we also provide a text HUD each step with last step’s numbers.
\emph{Trust the text HUD over pixels} for numeric state diagnosis (food/drink/energy/safety).
An \textbf{icon legend} image is attached at each step to resolve object appearance (player, terrain, resources, stations, creatures). Expect darker icons at night and be more conservative.

\medskip
\textbf{ENVIRONMENT \& OBJECTS (identify before acting).}
Water (lake) enables drinking; sand often borders lakes.
Trees/grass yield wood; stone and ores appear on rocky terrain (coal/iron/diamond require stronger tools).
Caves are darker stone; hostile creatures are more common at night. Lava is hazardous; bridge via placed stone.
Stations/placeables include table and furnace (for advanced tools), plus stone blocks and plants.

\medskip
\textbf{SURVIVAL-FIRST POLICY.}
When energy is low (e.g., $<6$), prioritize sleep. Night increases danger and reduces visibility.
If low water: seek the nearest lake and drink; if low food: harvest plants or obtain food from a cow.
Avoid action spamming; if repeated interactions have no effect, you are likely not adjacent or not facing the target.

\medskip
\textbf{PROGRESSION (only when needs are safe).}
Early: gather wood $\rightarrow$ place table $\rightarrow$ craft wood tools (including a defensive sword).
Mid: obtain stone/coal $\rightarrow$ place furnace $\rightarrow$ craft stone tools and mine iron.
Iron tier: with table+furnace and iron+coal+wood, craft iron tools and then mine diamond.
Keep stations in safe locations; use stone placement to bridge lava or block chokepoints; plant saplings for future food.

\medskip
\textbf{REASONING STYLE.}
Provide a brief diagnosis (health / hunger / thirst/ energy) and a concise next-step plan.
Be specific about your intended target (e.g., lake / tree / table/ cow) and your current facing direction.
\end{tcolorbox}

%% file: 9_appendix_template.tex
\section{Full Question Template}\label{app:questiontemplate}

\begin{table*}
  \centering
  \scriptsize
  \setlength{\tabcolsep}{3.3pt}
  \renewcommand{\arraystretch}{1.12}
  \begin{tabular}{p{0.11\textwidth} p{0.19\textwidth} p{0.56\textwidth} p{0.08\textwidth}}
    \toprule
    \textbf{Type} & \textbf{Template} & \textbf{Question template} & \textbf{Ans.} \\
    \midrule

    \textbf{Single-Hop} & \texttt{A\_action} &
    At step \{step\}, what action did you execute? &
    Action \\

    & \texttt{A\_reason} &
    At step \{step\}, what is the first sentence of your reasoning? &
    String \\

    & \texttt{A\_location} &
    Before performing the action at step \{step\}, what is your location? &
    Location \\

    & \texttt{A\_obs\_before} &
    Before performing the action at step \{step\}, what is the observation? &
    String \\

    & \texttt{A\_obs\_after} &
    After performing the action at step \{step\}, what is the resulting observation? &
    String \\

    & \texttt{A\_reward} &
    After performing the action at step \{step\}, what is the cumulative reward so far? &
    Integer \\

    & \texttt{A\_valid\_action} &
    At step \{step\}, is `\{action\}' a valid action? Answer in `yes' or `no'. &
    Yes/No \\

    & \texttt{A\_gain\_item} &
    At which step did you \{first / last\} gain `\{item\}'? &
    Step \# \\

    & \texttt{A\_enter\_leave} &
    At which step did you \{first start being at / first leave / last start being at\} `\{location\}' before performing the action? &
    Step \# \\

    & \texttt{A\_keyword\_occurrence} &
    Which step is the \{first / second / last / second last\} step whose \textbf{reason} mentions `\{keyword\}'? &
    Step \# \\

    \midrule
    \textbf{Multi-Hop} & \texttt{B\_gain\_after\_action} &
    After first obtaining `\{item\}', what action is executed \{delta\} step(s) later? &
    Action \\

    & \texttt{B\_gain\_after\_location} &
    After first obtaining `\{item\}', what is your location \{delta\} step(s) later? &
    Location \\

    & \texttt{B\_gain\_after\_observation} &
    After first obtaining `\{item\}', what is the observation \{delta\} step(s) later? &
    String \\

    & \texttt{B\_gain\_after\_reward} &
    After first obtaining `\{item\}', what is the cumulative reward \{delta\} step(s) later? &
    Integer \\

    & \texttt{B\_keyword\_after\_action} &
    After the first step whose reason mentions `\{keyword\}', what action is executed \{delta\} step(s) later? &
    Action \\

    & \texttt{B\_keyword\_after\_location} &
    After the first step whose reason mentions `\{keyword\}', what is your location \{delta\} step(s) later? &
    Location \\

    & \texttt{B\_keyword\_after\_\allowbreak observation} &
    After the first step whose reason mentions `\{keyword\}', what is the observation \{delta\} step(s) later? &
    String \\

    & \texttt{B\_keyword\_after\_reward} &
    After the first step whose reason mentions `\{keyword\}', what is the cumulative reward \{delta\} step(s) later? &
    Integer \\

    \midrule
    \textbf{Induction} & \texttt{C\_action\_mode} &
    What is the most frequent action executed? &
    Action \\

    & \texttt{C\_distinct\_locations} &
    From steps \{L\} to \{R\}, how many distinct locations were visited? &
    Integer \\

    & \texttt{C\_most\_frequent\_location} &
    From steps \{L\} to \{R\}, what is the most frequently visited location? &
    Location \\

    & \texttt{C\_total\_dwell} &
    Which is the location with the longest duration of stay in total (within \{L\}--\{R\})? &
    Location \\

    & \texttt{C\_keyword\_count\_obs} &
    From steps \{L\} to \{R\}, how many times does `\{keyword\}' appear in observations? &
    Integer \\

    & \texttt{C\_keyword\_count\_reason} &
    From steps \{L\} to \{R\}, how many times does `\{keyword\}' appear in reasons? &
    Integer \\

    \midrule
    \textbf{Spatial} & \texttt{D\_compare\_distances} &
    Which is closer (fewer steps to reach), `\{A\}' or `\{B\}', from the location at step \{anchor\}? &
    Location \\

    & \texttt{D\_direction\_count} &
    From steps \{L\} to \{R\}, how many times did you move \{direction\}? &
    Integer \\

    & \texttt{D\_reachable\_locations\_\allowbreak count} &
    How many distinct locations are reachable from `\{location\}' within \{k\} steps? &
    Integer \\

    & \texttt{D\_reachable\_within} &
    Is `\{target\}' reachable from `\{source\}' within \{k\} steps? Answer in `yes' or `no'. &
    Yes/No \\

    & \texttt{D\_sequence\_moves} &
    At step \{step\}, if you move \{sequence of directions\}, which location do you reach? &
    Location \\

    & \texttt{D\_shortest\_path} &
    What is the shortest path length between the locations at steps \{i\} and \{j\}? &
    Integer \\

    \midrule
    \textbf{Temporal} & \texttt{E\_gain\_delay} &
    After first obtaining `\{item\}', how many steps later is `\{item\}' obtained again? &
    Integer \\

    & \texttt{E\_item\_before\_leave} &
    Before first leaving `\{location\}', did you ever obtain `\{item\}'? Answer in `yes' or `no'. &
    Yes/No \\

    & \texttt{E\_item\_order} &
    Before first time obtaining `\{A\}', have you ever obtained `\{B\}'? Answer in `yes' or `no'. &
    Yes/No \\

    & \texttt{E\_region\_stay} &
    After first entering `\{location\}', for how many consecutive steps did you stay in `\{location\}'? &
    Integer \\

    & \texttt{E\_scene\_order} &
    Before first time entering `\{A\}', have you ever been to `\{B\}'? Answer in `yes' or `no'. &
    Yes/No \\

    \midrule
    \textbf{Logical} & \texttt{F\_has\_item} &
    At step \{step\}, did you have `\{item\}' in inventory? Answer in `yes' or `no'. &
    Yes/No \\

    & \texttt{F\_list\_inventory} &
    After performing the action at step \{step\}, what are all the items you carry? &
    List \\

    & \texttt{F\_max\_inventory\_step} &
    After completing which step does the inventory contain the most items? &
    Step \# \\

    & \texttt{F\_location\_most\_\allowbreak item\_gain} &
    At which location did you gain the most distinct items? &
    Location \\

    \midrule
    \textbf{Adversarial} & \texttt{---} &
    \textbf{Construction rule:} take any non-adversarial template above and instantiate it with a value that does \emph{not} occur in the trajectory but valid in the game environment (e.g., item/location/keyword), creating a false-premise question. The correct response should be \texttt{not answerable}. &
    Not Answerable\\

    &  &
    \textbf{Example:} instantiate \texttt{A\_gain\_item} with an unseen item.\newline
    ``At which step did you first gain `\texttt{sword}'?'' $\rightarrow$ \texttt{not answerable}. &
     \\

    \bottomrule
  \end{tabular}
  \caption{Full question template inventory for Jericho text games. Keyword-occurrence variants are merged; Multi-Hop templates are listed individually. Adversarial templates are constructed by instantiating existing templates with nonexistent (false-premise) values; see example.}
  \label{tab:jericho_full_question_templates}
\end{table*}


\begin{table*}
  \centering
  \scriptsize
  \setlength{\tabcolsep}{3.3pt}
  \renewcommand{\arraystretch}{1.12}
  \begin{tabular}{p{0.11\textwidth} p{0.19\textwidth} p{0.56\textwidth} p{0.08\textwidth}}
    \toprule
    \textbf{Type} & \textbf{Template} & \textbf{Question template} & \textbf{Ans.} \\
    \midrule

    \textbf{Single-Hop} & \texttt{A\_action} &
    What is the action at step \{step\}? &
    Action \\

    & \texttt{A\_reason} &
    What was the reasoning at step \{step\}? &
    String \\

    & \texttt{A\_stat} &
    What was your \{stat\} value at step \{step\}? &
    Integer \\

    & \texttt{A\_terrain} &
    What terrain was under you at step \{step\}? &
    Terrain \\

    & \texttt{A\_inventory} &
    How many \{resource\} did you have at step \{step\}? &
    Integer \\

    & \texttt{A\_occ\_action} &
    Which step is the \{first / second / third / last\} step whose action is `\{action\}'? &
    Step \# \\

    & \texttt{A\_occ\_keyword} &
    Which step is the \{first / second / third / last\} step whose reason mentions `\{keyword\}'? &
    Step \# \\

    & \texttt{A\_occ\_terrain} &
    Which step is the \{first / second / third / last\} step which you stood on \{terrain\} terrain? &
    Step \# \\

    \midrule
    \textbf{Multi-Hop} & \texttt{B\_action} &
    What is the action \{offset\} step(s) \{before / after\} the \{first / second / third / last\} step whose action is `\{action\}'? &
    Action \\

    & \texttt{B\_keyword} &
    What is the action \{offset\} step(s) \{before / after\} the \{first / second / third / last\} step whose reason mentions `\{keyword\}'? &
    Action \\

    & \texttt{B\_terrain} &
    What is the action \{offset\} step(s) \{before / after\} the \{first / second / third / last\} step which you stood on \{terrain\} terrain? &
    Action \\

    \midrule
    \textbf{Induction} & \texttt{C\_keyword\_count} &
    From steps \{L\} to \{R\}, how many steps have reasons that mention `\{keyword\}'? &
    Integer \\

    & \texttt{C\_most\_move} &
    From steps \{L\} to \{R\}, what was the most common movement direction? &
    Direction \\

    & \texttt{C\_longest\_run} &
    From steps \{L\} to \{R\}, what was the longest consecutive run of \{action\}? &
    Integer \\

    & \texttt{C\_collect\_res} &
    From steps \{L\} to \{R\}, how many times did you successfully collect \{resource\}? &
    Integer \\

    & \texttt{C\_resource\_peak} &
    After completing which step does \{resource\} reach its maximum quantity (at least 1)? &
    Step \# \\

    & \texttt{C\_resource\_change} &
    From steps \{L\} to \{R\}, what was the change in \{resource\} quantity? &
    Integer \\

    & \texttt{C\_visible\_count} &
    From steps \{L\} to \{R\}, in how many steps could you see any \{terrain\} in the frame? &
    Integer \\

    & \texttt{C\_adjacent\_count} &
    From steps \{L\} to \{R\}, in how many steps were you adjacent to any \{terrain\}? &
    Integer \\

    & \texttt{C\_distinct\_trees} &
    From steps \{L\} to \{R\}, how many distinct trees did you see in total? &
    Integer \\

    \midrule
    \textbf{Spatial} & \texttt{D\_displacement} &
    From steps \{L\} to \{R\}, what was the total displacement of you?
    Answer in `X step(s) left/right and Y step(s) up/down'. &
    Dis\-place\-ment \\

    & \texttt{D\_path\_length} &
    From steps \{L\} to \{R\}, how many movement steps did you successfully take? &
    Integer \\

    & \texttt{D\_predict\_terrain} &
    At step \{step\}, if you walked \{direction\} \{K\} step(s), what terrain would be underfoot? &
    Terrain \\

    & \texttt{D\_nearest\_dir} &
    At step \{step\}, in which direction is the nearest \{terrain\} relative to you? &
    Direction \\

    & \texttt{D\_nav\_to\_target} &
    At step \{step\}, how should you move to reach the nearest \{terrain\}?
    Answer in displacement format (or `0' if already on target). &
    Dis\-place\-ment \\

    & \texttt{D\_min\_dist} &
    From steps \{L\} to \{R\}, at which step were you closest to the nearest \{terrain\}? (If ties, answer the first.) &
    Step \# \\

    & \texttt{D\_max\_dist} &
    From steps \{L\} to \{R\}, at which step were you furthest from the nearest \{terrain\}? (If ties, answer the first.) &
    Step \# \\

    \midrule
    \textbf{Temporal} & \texttt{E\_event\_order} &
    Did \{eventA\} happen before \{eventB\}? Answer in `yes' or `no'. &
    Yes/No \\

    & \texttt{E\_event\_interval} &
    After \{eventA\}, after how many steps did \{eventB\} occur? &
    Integer \\

    & \texttt{E\_state\_after\_event} &
    Immediately after \{event\}, what was your \{stat\} value? &
    Integer \\

    \midrule
    \textbf{Logical} & \texttt{F\_craft\_feasibility} &
    At step \{step\}, are the collected resources enough to craft \{craft\}? Answer in `yes' or `no'. &
    Yes/No \\

    & \texttt{F\_event\_loc} &
    At which step(s) did you \{event\}? (Answer as a list of step numbers.) &
    List \\

    & \texttt{F\_attack\_count} &
    How many times were you attacked? &
    Integer \\

    & \texttt{F\_death\_reason} &
    What was the cause of your death at the end of the episode? &
    String \\

    & \texttt{F\_first\_attack\_step} &
    At which step were you attacked for the first time? &
    Step \# \\

    & \texttt{F\_last\_attack\_step} &
    At which step were you attacked for the last time? &
    Step \# \\

    & \texttt{F\_inventory\_contents} &
    At step \{step\}, what are all the items you carry? &
    List \\

    \midrule
    \textbf{Adversarial} & \texttt{---} &
    \textbf{Construction rule:} take any non-adversarial template above and instantiate it with a value that does \emph{not} occur in the trajectory but valid in the game environment (e.g., resource/terrain/keyword/action), creating a false-premise question. The correct response should be \texttt{not answerable}. &
    Not Answerable\\

    &  &
    \textbf{Example:} instantiate \texttt{A\_occ\_keyword} with an unseen keyword.\newline
    ``Which step is the first step whose reason mentions `\texttt{teleport}'?'' $\rightarrow$ \texttt{not answerable}. &
     \\

    \bottomrule
  \end{tabular}
  \caption{Full question template inventory for the visual-game (Crafter) benchmark. Occurrence variants are merged; Multi-Hop templates are listed individually.}
  \label{tab:crafter_full_question_templates}
\end{table*}

We provide the full question template in Table~\ref{tab:jericho_full_question_templates} for text games and in Table~\ref{tab:crafter_full_question_templates} for visual games.

%% file: 9_appendix_algorithm.tex
\section{Demonstration of Ground Truth Computation Algorithm}
\label{app:gt_d_nav_to_target}
We demonstrate the ground-truth computation using  \texttt{D\_nav\_to\_target} questions as an example.

We compute its ground-truth by (i) building the dynamic grid for the episode,
(ii) running BFS from the query step position to all terrain cells matching the target terrain, and
(iii) reconstructing one or multiple shortest move sequences. See Algorithm~\ref{alg:dn_nav_to_target} and Algorithm~\ref{alg:bfs_all_min_targets}.

\begin{algorithm}
\caption{Ground-truth for \texttt{D\_nav\_to\_target}}
\label{alg:dn_nav_to_target}
\begin{algorithmic}[1]\raggedright
\Require base grid $G$; trajectory positions $\texttt{pos}[t]=(r_t,c_t)$; actions $\texttt{act}[t]$; inventories $\texttt{inv}[t]$; optional dynamic maps $\texttt{maps}[t]$.
\Require query step $t_0$ (0-based internally); target terrain name $\tau$.
\Ensure ground-truth answer $y$ (\textsc{String} / \textsc{List} / \texttt{not answerable}).

\State \textit{/* Build / select dynamic grid */}
\If{$\texttt{maps}$ is available}
  \State $\texttt{dyn\_grid} \gets \texttt{maps}[\texttt{last}]$
\Else
  \State \texttt{dyn\_grid} $\gets$ \textsc{ApplyTrajectoryEdits}($G$, \texttt{pos}, \texttt{act},
  \StatexIndent[2] \texttt{inv})
\EndIf

\State $\texttt{start} \gets \texttt{pos}[t_0]$
\State $\texttt{target\_ids} \gets \{\, id \mid \textsc{TerrainIdToName}[id] = \tau \,\}$

\State $(d,\ \texttt{targets},\ \texttt{parent}) \gets$ \textsc{BFSAllMinTargets}(\texttt{dyn\_grid}, \texttt{start},
\StatexIndent \texttt{target\_ids}) \Comment{Algorithm~\ref{alg:bfs_all_min_targets}}

\If{$d=\texttt{None}$}
  \State \Return \texttt{not answerable}
\ElsIf{$d=0$}
  \State \Return $0$
\EndIf

\State $\texttt{routes} \gets \emptyset$
\ForAll{$u \in \texttt{targets}$}
  \State $\texttt{dirs} \gets$ \textsc{ReconstructPath}(\texttt{parent}, \texttt{start}, $u$)
  \State $\texttt{routes} \gets \texttt{routes} \cup \{\,\textsc{CompressPath}(\texttt{dirs})\,\}$
\EndFor

\If{$|\texttt{routes}|=0$} \Comment{safeguard}
  \State \Return \texttt{not answerable}
\ElsIf{$|\texttt{routes}|=1$}
  \State \Return the only element in \texttt{routes}
\Else
  \State \Return \textsc{SortedList}(\texttt{routes})
\EndIf
\end{algorithmic}
\end{algorithm}

\begin{algorithm}
\caption{\textsc{BFSAllMinTargets}}
\label{alg:bfs_all_min_targets}
\begin{algorithmic}[1]\raggedright
\Require grid \texttt{dyn\_grid}; start cell \texttt{start}; target terrain id set \texttt{target\_ids}.
\Ensure $(d,\ \texttt{targets},\ \texttt{parent})$, where $d$ is the shortest distance to any target terrain cell; \texttt{targets} are all target cells at distance $d$; \texttt{parent} stores BFS tree for path reconstruction.

\State initialize FIFO queue $Q \gets [\texttt{start}]$
\State initialize distance map \texttt{dist[start]} $\gets 0$
\State initialize parent map \texttt{parent} (empty)

\State $\texttt{best\_d} \gets \texttt{None}$, $\texttt{targets} \gets \emptyset$
\While{$Q$ not empty}
  \State pop cell $x$ from $Q$
  \State $d_x \gets \texttt{dist}[x]$

  \If{$\texttt{best\_d} \neq \texttt{None}$ \textbf{and} $d_x > \texttt{best\_d}$}
    \State \textbf{break} \Comment{all remaining nodes are farther}
  \EndIf

  \If{\textsc{TerrainId}(\texttt{dyn\_grid}, $x$) $\in \texttt{target\_ids}$}
    \If{$\texttt{best\_d}=\texttt{None}$}
      \State $\texttt{best\_d} \gets d_x$
    \EndIf
    \State $\texttt{targets} \gets \texttt{targets} \cup \{x\}$
    \State \textbf{continue}
  \EndIf

  \ForAll{4-neighbor $y$ of $x$ that is traversable}
    \If{$y$ not visited}
      \State \texttt{dist[$y$]} $\gets d_x + 1$
      \State \texttt{parent[$y$]} $\gets x$
      \State push $y$ into $Q$
    \EndIf
  \EndFor
\EndWhile

\If{$\texttt{best\_d}=\texttt{None}$}
  \State \Return $(\texttt{None},\ \emptyset,\ \texttt{parent})$
\Else
  \State \Return $(\texttt{best\_d},\ \texttt{targets},\ \texttt{parent})$
\EndIf
\end{algorithmic}
\end{algorithm}

%% file: custom.bib
@article{survey_agent,
  author       = {Lei Wang and
                  Chen Ma and
                  Xueyang Feng and
                  Zeyu Zhang and
                  Hao Yang and
                  Jingsen Zhang and
                  Zhiyuan Chen and
                  Jiakai Tang and
                  Xu Chen and
                  Yankai Lin and
                  Wayne Xin Zhao and
                  Zhewei Wei and
                  Jirong Wen},
  title        = {A survey on large language model based autonomous agents},
  journal      = {Frontiers Comput. Sci.},
  volume       = {18},
  number       = {6},
  pages        = {186345},
  year         = {2024},
  url          = {https://doi.org/10.1007/s11704-024-40231-1},
  doi          = {10.1007/S11704-024-40231-1},
  bibsource    = {dblp computer science bibliography, https://dblp.org}
}

@inproceedings{brown2020language,
  author       = {Tom B. Brown and
                      Benjamin Mann and
                      Nick Ryder and
                      Melanie Subbiah and
                      Jared Kaplan and
                      Prafulla Dhariwal and
                      Arvind Neelakantan and
                      Pranav Shyam and
                      Girish Sastry and
                      Amanda Askell and
                      Sandhini Agarwal and
                      Ariel Herbert{-}Voss and
                      Gretchen Krueger and
                      Tom Henighan and
                      Rewon Child and
                      Aditya Ramesh and
                      Daniel M. Ziegler and
                      Jeffrey Wu and
                      Clemens Winter and
                      Christopher Hesse and
                      Mark Chen and
                      Eric Sigler and
                      Mateusz Litwin and
                      Scott Gray and
                      Benjamin Chess and
                      Jack Clark and
                      Christopher Berner and
                      Sam McCandlish and
                      Alec Radford and
                      Ilya Sutskever and
                      Dario Amodei},
  editor       = {Hugo Larochelle and
                  Marc'Aurelio Ranzato and
                  Raia Hadsell and
                  Maria{-}Florina Balcan and
                  Hsuan{-}Tien Lin},
  title        = {Language Models are Few-Shot Learners},
  booktitle    = {Advances in Neural Information Processing Systems 33: Annual Conference
                  on Neural Information Processing Systems 2020, NeurIPS 2020, December
                  6-12, 2020, virtual},
  year         = {2020},
  pages        = {1877--1901},
  url          = {https://proceedings.neurips.cc/paper/2020/hash/1457c0d6bfcb4967418bfb8ac142f64a-Abstract.html},
  bibsource    = {dblp computer science bibliography, https://dblp.org}
}

@inproceedings{locomo,
  author       = {Adyasha Maharana and
                  Dong{-}Ho Lee and
                  Sergey Tulyakov and
                  Mohit Bansal and
                  Francesco Barbieri and
                  Yuwei Fang},
  editor       = {Lun{-}Wei Ku and
                  Andre Martins and
                  Vivek Srikumar},
  title        = {Evaluating Very Long-Term Conversational Memory of {LLM} Agents},
  booktitle    = {Proceedings of the 62nd Annual Meeting of the Association for Computational
                  Linguistics (Volume 1: Long Papers), {ACL} 2024, Bangkok, Thailand,
                  August 11-16, 2024},
  pages        = {13851--13870},
  publisher    = {Association for Computational Linguistics},
  year         = {2024},
  url          = {https://doi.org/10.18653/v1/2024.acl-long.747},
  doi          = {10.18653/V1/2024.ACL-LONG.747},
  bibsource    = {dblp computer science bibliography, https://dblp.org}
}

@article{survey_memory,
  author       = {Zeyu Zhang and
                  Quanyu Dai and
                  Xiaohe Bo and
                  Chen Ma and
                  Rui Li and
                  Xu Chen and
                  Jieming Zhu and
                  Zhenhua Dong and
                  Ji{-}Rong Wen},
  title        = {A Survey on the Memory Mechanism of Large Language Model-based Agents},
  journal      = {{ACM} Trans. Inf. Syst.},
  volume       = {43},
  number       = {6},
  pages        = {155:1--155:47},
  year         = {2025},
  url          = {https://doi.org/10.1145/3748302},
  doi          = {10.1145/3748302},
  bibsource    = {dblp computer science bibliography, https://dblp.org}
}

@article{lc_vs_rag,
  author       = {Xinze Li and
                  Yixin Cao and
                  Yubo Ma and
                  Aixin Sun},
  title        = {Long Context vs. {RAG} for {LLMs}: An Evaluation and Revisits},
  journal      = {CoRR},
  volume       = {abs/2501.01880},
  year         = {2025},
  url          = {https://doi.org/10.48550/arXiv.2501.01880},
  doi          = {10.48550/ARXIV.2501.01880},
  eprinttype    = {arXiv},
  eprint       = {2501.01880},
  bibsource    = {dblp computer science bibliography, https://dblp.org}
}

@inproceedings{longmemeval,
  author       = {Di Wu and
                  Hongwei Wang and
                  Wenhao Yu and
                  Yuwei Zhang and
                  Kai{-}Wei Chang and
                  Dong Yu},
  title        = {{LongMemEval}: Benchmarking Chat Assistants on Long-Term Interactive
                  Memory},
  booktitle    = {The Thirteenth International Conference on Learning Representations,
                  {ICLR} 2025, Singapore, April 24-28, 2025},
  publisher    = {OpenReview.net},
  year         = {2025},
  url          = {https://openreview.net/forum?id=pZiyCaVuti},
  bibsource    = {dblp computer science bibliography, https://dblp.org}
}

@article{melton1963memory,
  author  = {Melton, Arthur W.},
  title   = {Implications of Short-Term Memory for a General Theory of Memory},
  journal = {Journal of Verbal Learning and Verbal Behavior},
  year    = {1963},
  volume  = {2},
  number  = {1},
  pages   = {1--21},
  doi     = {10.1016/S0022-5371(63)80063-8}
}

@book{matlin2005cognition,
  author    = {Matlin, Margaret W.},
  title     = {Cognition},
  edition   = {6},
  year      = {2005},
  publisher = {John Wiley \& Sons},
  address   = {Hoboken, NJ}
}

@book{sternberg1999cognitive,
  author    = {Sternberg, Robert J.},
  title     = {Cognitive Psychology},
  edition   = {2},
  year      = {1999},
  publisher = {Harcourt Brace College Publishers},
  address   = {Fort Worth, TX}
}

@article{greenberg2010interdependence,
  author  = {Greenberg, Daniel L. and Verfaellie, Mieke},
  title   = {Interdependence of Episodic and Semantic Memory: Evidence from Neuropsychology},
  journal = {Journal of the International Neuropsychological Society},
  year    = {2010},
  volume  = {16},
  number  = {5},
  pages   = {748--753},
  doi     = {10.1017/S1355617710000676}
}

@article{tulving1973encoding,
  author  = {Tulving, Endel and Thomson, Donald M.},
  title   = {Encoding Specificity and Retrieval Processes in Episodic Memory},
  journal = {Psychological Review},
  year    = {1973},
  volume  = {80},
  number  = {5},
  pages   = {352--373},
  doi     = {10.1037/h0020071}
}

@article{lostin,
  author       = {Nelson F. Liu and
                  Kevin Lin and
                  John Hewitt and
                  Ashwin Paranjape and
                  Michele Bevilacqua and
                  Fabio Petroni and
                  Percy Liang},
  title        = {Lost in the Middle: How Language Models Use Long Contexts},
  journal      = {Trans. Assoc. Comput. Linguistics},
  volume       = {12},
  pages        = {157--173},
  year         = {2024},
  url          = {https://doi.org/10.1162/tacl\_a\_00638},
  doi          = {10.1162/TACL\_A\_00638},
  bibsource    = {dblp computer science bibliography, https://dblp.org}
}

@inproceedings{dpr,
  author       = {Vladimir Karpukhin and
                  Barlas Oguz and
                  Sewon Min and
                  Patrick Lewis and
                  Ledell Wu and
                  Sergey Edunov and
                  Danqi Chen and
                  Wen{-}tau Yih},
  editor       = {Bonnie Webber and
                  Trevor Cohn and
                  Yulan He and
                  Yang Liu},
  title        = {Dense Passage Retrieval for Open-Domain Question Answering},
  booktitle    = {Proceedings of the 2020 Conference on Empirical Methods in Natural
                  Language Processing, {EMNLP} 2020, Online, November 16-20, 2020},
  pages        = {6769--6781},
  publisher    = {Association for Computational Linguistics},
  year         = {2020},
  url          = {https://doi.org/10.18653/v1/2020.emnlp-main.550},
  doi          = {10.18653/V1/2020.EMNLP-MAIN.550},
  bibsource    = {dblp computer science bibliography, https://dblp.org}
}

@inproceedings{ragfornlptasks,
  author       = {Patrick Lewis and
                  Ethan Perez and
                  Aleksandra Piktus and
                  Fabio Petroni and
                  Vladimir Karpukhin and
                  Naman Goyal and
                  Heinrich K{\"{u}}ttler and
                  Mike Lewis and
                  Wen{-}tau Yih and
                  Tim Rockt{\"{a}}schel and
                  Sebastian Riedel and
                  Douwe Kiela},
  editor       = {Hugo Larochelle and
                  Marc'Aurelio Ranzato and
                  Raia Hadsell and
                  Maria{-}Florina Balcan and
                  Hsuan{-}Tien Lin},
  title        = {Retrieval-Augmented Generation for Knowledge-Intensive {NLP} Tasks},
  booktitle    = {Advances in Neural Information Processing Systems 33: Annual Conference
                  on Neural Information Processing Systems 2020, NeurIPS 2020, December
                  6-12, 2020, virtual},
  year         = {2020},
  pages        = {9459--9474},
  url          = {https://proceedings.neurips.cc/paper/2020/hash/6b493230205f780e1bc26945df7481e5-Abstract.html},
  bibsource    = {dblp computer science bibliography, https://dblp.org}
}

@inproceedings{mem0,
  title     = {{Mem0}: Building Production-Ready {AI} Agents with Scalable Long-Term Memory},
  author    = {Chhikara, Prateek and Khant, Dev and Aryan, Saket and Singh, Taranjeet and Yadav, Deshraj},
  booktitle = {Proceedings of the 28th European Conference on Artificial Intelligence (ECAI 2025)},
  publisher = {IOS Press},
  year      = {2025},
  doi       = {10.3233/FAIA251160},
  pages     = {2993--3000}
}

@article{langmem,
  title        = {{LangMem}: Long-term Memory for {LLM} Agents},
  author       = {{LangChain}},
  year         = {2025},
  url          = {https://github.com/langchain-ai/langmem},
  note         = {Accessed: 2025-12-28}
}

@article{memgpt,
  author       = {Charles Packer and Sarah Wooders and Kevin Lin and Vivian Fang and Shishir G. Patil and Ion Stoica and Joseph E. Gonzalez},
  title        = {{MemGPT}: Towards {LLMs} as Operating Systems},
  journal      = {CoRR},
  volume       = {abs/2310.08560},
  year         = {2023},
  url          = {https://doi.org/10.48550/arXiv.2310.08560},
  doi          = {10.48550/ARXIV.2310.08560},
  eprinttype    = {arXiv},
  eprint       = {2310.08560},
  bibsource    = {dblp computer science bibliography, https://dblp.org}
}

@inproceedings{amem,
  title     = {{A-Mem}: Agentic Memory for {LLM} Agents},
  author    = {Xu, Wujiang and Liang, Zujie and Mei, Kai and Gao, Hang and Tan, Juntao and Zhang, Yongfeng},
  booktitle = {Advances in Neural Information Processing Systems 38},
  year      = {2025},
  doi       = {10.52202/085713-0593},
  pages     = {17577--17604}
}

@inproceedings{memagent,
  title     = {{MemAgent}: Reshaping Long-Context {LLM} with Multi-Conv {RL}-based Memory
                  Agent},
  author    = {Hongli Yu and
                  Tinghong Chen and
                  Jiangtao Feng and
                  Jiangjie Chen and
                  Weinan Dai and
                  Qiying Yu and
                  Ya{-}Qin Zhang and
                  Wei{-}Ying Ma and
                  Jingjing Liu and
                  Mingxuan Wang and
                  Hao Zhou},
  booktitle = {The Fourteenth International Conference on Learning Representations (ICLR)},
  year      = {2026},
  pages     = {39458--39486},
  url       = {https://proceedings.iclr.cc/paper_files/paper/2026/hash/4264ee4376776907c0b87ed70b959585-Abstract-Conference.html}
}

@inproceedings{memory-r1,
  title     = {Memory-{R1}: Enhancing Large Language Model Agents to Manage and Utilize Memories via Reinforcement Learning},
  author    = {Yan, Sikuan and Yang, Xiufeng and Huang, Zuchao and Nie, Ercong and Ding, Zifeng and Li, Zonggen and Ma, Xiaowen and Bi, Jinhe and Kersting, Kristian and Pan, Jeff Z. and Schuetze, Hinrich and Tresp, Volker and Ma, Yunpu},
  booktitle = {Proceedings of the 64th Annual Meeting of the Association for Computational Linguistics (Volume 1: Long Papers)},
  year      = {2026},
  address   = {San Diego, California, USA},
  publisher = {Association for Computational Linguistics},
  url       = {https://aclanthology.org/2026.acl-long.583/},
  doi       = {10.18653/v1/2026.acl-long.583},
  pages     = {12805--12825}
}

@article{mema,
  author       = {Yu Wang and
                  Ryuichi Takanobu and
                  Zhiqi Liang and
                  Yuzhen Mao and
                  Yuanzhe Hu and
                  Julian J. McAuley and
                  Xiaojian Wu},
  title        = {{Mem-\(\alpha\)}: Learning Memory Construction via Reinforcement Learning},
  journal      = {CoRR},
  volume       = {abs/2509.25911},
  year         = {2025},
  url          = {https://doi.org/10.48550/arXiv.2509.25911},
  doi          = {10.48550/ARXIV.2509.25911},
  eprinttype    = {arXiv},
  eprint       = {2509.25911},
  bibsource    = {dblp computer science bibliography, https://dblp.org}
}

@inproceedings{MemoryBank,
  author       = {Wanjun Zhong and
                  Lianghong Guo and
                  Qiqi Gao and
                  He Ye and
                  Yanlin Wang},
  editor       = {Michael J. Wooldridge and
                  Jennifer G. Dy and
                  Sriraam Natarajan},
  title        = {{MemoryBank}: Enhancing Large Language Models with Long-Term Memory},
  booktitle    = {Thirty-Eighth {AAAI} Conference on Artificial Intelligence, {AAAI}
                  2024, Thirty-Sixth Conference on Innovative Applications of Artificial
                  Intelligence, {IAAI} 2024, Fourteenth Symposium on Educational Advances
                  in Artificial Intelligence, {EAAI} 2014, February 20-27, 2024, Vancouver,
                  Canada},
  pages        = {19724--19731},
  publisher    = {{AAAI} Press},
  year         = {2024},
  url          = {https://doi.org/10.1609/aaai.v38i17.29946},
  doi          = {10.1609/AAAI.V38I17.29946},
  bibsource    = {dblp computer science bibliography, https://dblp.org}
}

@inproceedings{PerLTQA,
  title     = {{P}er{LTQA}: A Personal Long-Term Memory Dataset for Memory Classification, Retrieval, and Fusion in Question Answering},
  author    = {Du, Yiming and Wang, Hongru and Zhao, Zhengyi and Liang, Bin and Wang, Baojun and Zhong, Wanjun and Wang, Zezhong and Wong, Kam-Fai},
  booktitle = {Proceedings of the 10th SIGHAN Workshop on Chinese Language Processing (SIGHAN-10)},
  year      = {2024},
  address   = {Bangkok, Thailand},
  publisher = {Association for Computational Linguistics},
  url       = {https://aclanthology.org/2024.sighan-1.18/},
  pages     = {152--164}
}

@inproceedings{beam,
  title     = {Beyond a Million Tokens: Benchmarking and Enhancing Long-Term Memory
                  in {LLMs}},
  author    = {Mohammad Tavakoli and
                  Alireza Salemi and
                  Carrie Ye and
                  Mohamed Abdalla and
                  Hamed Zamani and
                  J. Ross Mitchell},
  booktitle = {The Fourteenth International Conference on Learning Representations (ICLR)},
  year      = {2026},
  pages     = {133343--133429},
  url       = {https://proceedings.iclr.cc/paper_files/paper/2026/hash/d7f0cfa0fe759b033d5262e1bb7d4065-Abstract-Conference.html}
}

@inproceedings{membench,
  title     = {{M}em{B}ench: Towards More Comprehensive Evaluation on the Memory of {LLM}-based Agents},
  author    = {Tan, Haoran and Zhang, Zeyu and Ma, Chen and Chen, Xu and Dai, Quanyu and Dong, Zhenhua},
  booktitle = {Findings of the Association for Computational Linguistics: ACL 2025},
  year      = {2025},
  address   = {Vienna, Austria},
  publisher = {Association for Computational Linguistics},
  url       = {https://aclanthology.org/2025.findings-acl.989/},
  doi       = {10.18653/v1/2025.findings-acl.989},
  pages     = {19336--19352}
}

@inproceedings{MemoryAgentBench,
  title     = {Evaluating Memory in {LLM} Agents via Incremental Multi-Turn Interactions},
  author    = {Yuanzhe Hu and
                  Yu Wang and
                  Julian McAuley},
  booktitle = {The Fourteenth International Conference on Learning Representations (ICLR)},
  year      = {2026},
  pages     = {156259--156291},
  url       = {https://proceedings.iclr.cc/paper_files/paper/2026/hash/fd1eff9dd295df50a41f2521942fa31d-Abstract-Conference.html}
}

@inproceedings{MemoryBench,
  author    = {Qingyao Ai and
               Yichen Tang and
               Changyue Wang and
               Jianming Long and
               Weihang Su and
               Yiqun Liu},
  title     = {{MemoryBench}: {A} Benchmark for Memory and Continual Learning in {LLM} Systems},
  booktitle = {Proceedings of the 43rd International Conference on Machine Learning (ICML)},
  year      = {2026},
  url       = {https://icml.cc/virtual/2026/poster/64918}
}

@article{StoryBench,
  author       = {Luanbo Wan and
                  Weizhi Ma},
  title        = {{StoryBench}: {A} Dynamic Benchmark for Evaluating Long-Term Memory
                  with Multi Turns},
  journal      = {CoRR},
  volume       = {abs/2506.13356},
  year         = {2025},
  url          = {https://doi.org/10.48550/arXiv.2506.13356},
  doi          = {10.48550/ARXIV.2506.13356},
  eprinttype    = {arXiv},
  eprint       = {2506.13356},
  bibsource    = {dblp computer science bibliography, https://dblp.org}
}

@inproceedings{jericho,
  author       = {Matthew J. Hausknecht and
                  Prithviraj Ammanabrolu and
                  Marc{-}Alexandre C{\^{o}}t{\'{e}} and
                  Xingdi Yuan},
  title        = {Interactive Fiction Games: {A} Colossal Adventure},
  booktitle    = {The Thirty-Fourth {AAAI} Conference on Artificial Intelligence, {AAAI}
                  2020, The Thirty-Second Innovative Applications of Artificial Intelligence
                  Conference, {IAAI} 2020, The Tenth {AAAI} Symposium on Educational
                  Advances in Artificial Intelligence, {EAAI} 2020, New York, NY, USA,
                  February 7-12, 2020},
  pages        = {7903--7910},
  publisher    = {{AAAI} Press},
  year         = {2020},
  url          = {https://doi.org/10.1609/aaai.v34i05.6297},
  doi          = {10.1609/AAAI.V34I05.6297},
  bibsource    = {dblp computer science bibliography, https://dblp.org}
}

@inproceedings{hafner2021crafter,
  author       = {Danijar Hafner},
  title        = {Benchmarking the Spectrum of Agent Capabilities},
  booktitle    = {The Tenth International Conference on Learning Representations, {ICLR}
                  2022, Virtual Event, April 25-29, 2022},
  publisher    = {OpenReview.net},
  year         = {2022},
  url          = {https://openreview.net/forum?id=1W0z96MFEoH},
  bibsource    = {dblp computer science bibliography, https://dblp.org}
}

@inproceedings{spring2023,
  title     = {{SPRING}: Studying Papers and Reasoning to Play Games},
  author    = {Wu, Yue and Min, So Yeon and Prabhumoye, Shrimai and Bisk, Yonatan and Salakhutdinov, Russ R. and Azaria, Amos and Mitchell, Tom M. and Li, Yuanzhi},
  booktitle = {Advances in Neural Information Processing Systems 36},
  year      = {2023},
  doi       = {10.52202/075280-0983},
  pages     = {22383--22687}
}

@inproceedings{rajpurkar2016squad,
  author       = {Pranav Rajpurkar and
                  Jian Zhang and
                  Konstantin Lopyrev and
                  Percy Liang},
  editor       = {Jian Su and
                  Xavier Carreras and
                  Kevin Duh},
  title        = {{SQuAD}: 100,000+ Questions for Machine Comprehension of Text},
  booktitle    = {Proceedings of the 2016 Conference on Empirical Methods in Natural
                  Language Processing, {EMNLP} 2016, Austin, Texas, USA, November 1-4,
                  2016},
  pages        = {2383--2392},
  publisher    = {The Association for Computational Linguistics},
  year         = {2016},
  url          = {https://doi.org/10.18653/v1/d16-1264},
  doi          = {10.18653/V1/D16-1264},
  bibsource    = {dblp computer science bibliography, https://dblp.org}
}

@inproceedings{biten2019stvqa,
  author       = {Ali Furkan Biten and
                  Rub{\`{e}}n Tito and
                  Andr{\'{e}}s Mafla and
                  Llu{\'{\i}}s G{\'{o}}mez i Bigorda and
                  Mar{\c{c}}al Rusi{\~{n}}ol and
                  C. V. Jawahar and
                  Ernest Valveny and
                  Dimosthenis Karatzas},
  title        = {Scene Text Visual Question Answering},
  booktitle    = {2019 {IEEE/CVF} International Conference on Computer Vision, {ICCV}
                  2019, Seoul, Korea (South), October 27 - November 2, 2019},
  pages        = {4290--4300},
  publisher    = {{IEEE}},
  year         = {2019},
  url          = {https://doi.org/10.1109/ICCV.2019.00439},
  doi          = {10.1109/ICCV.2019.00439},
  bibsource    = {dblp computer science bibliography, https://dblp.org}
}

@inproceedings{vanlandeghem2023dude,
  author       = {Jordy Van Landeghem and
                  Rafal Powalski and
                  Rub{\`{e}}n Tito and
                  Dawid Jurkiewicz and
                  Matthew B. Blaschko and
                  Lukasz Borchmann and
                  Micka{\"{e}}l Coustaty and
                  Sien Moens and
                  Michal Pietruszka and
                  Bertrand Anckaert and
                  Tomasz Stanislawek and
                  Pawel J{\'{o}}ziak and
                  Ernest Valveny},
  title        = {Document Understanding Dataset and Evaluation {(DUDE)}},
  booktitle    = {{IEEE/CVF} International Conference on Computer Vision, {ICCV} 2023,
                  Paris, France, October 1-6, 2023},
  pages        = {19471--19483},
  publisher    = {{IEEE}},
  year         = {2023},
  url          = {https://doi.org/10.1109/ICCV51070.2023.01789},
  doi          = {10.1109/ICCV51070.2023.01789},
  bibsource    = {dblp computer science bibliography, https://dblp.org}
}

@inproceedings{mmlongbenchdoc,
  author       = {Yubo Ma and
                  Yuhang Zang and
                  Liangyu Chen and
                  Meiqi Chen and
                  Yizhu Jiao and
                  Xinze Li and
                  Xinyuan Lu and
                  Ziyu Liu and
                  Yan Ma and
                  Xiaoyi Dong and
                  Pan Zhang and
                  Liangming Pan and
                  Yu{-}Gang Jiang and
                  Jiaqi Wang and
                  Yixin Cao and
                  Aixin Sun},
  editor       = {Amir Globerson and
                  Lester Mackey and
                  Danielle Belgrave and
                  Angela Fan and
                  Ulrich Paquet and
                  Jakub M. Tomczak and
                  Cheng Zhang},
  title        = {{MMLONGBENCH-DOC:} Benchmarking Long-context Document Understanding
                  with Visualizations},
  booktitle    = {Advances in Neural Information Processing Systems 37: Annual Conference
                  on Neural Information Processing Systems 2024, NeurIPS 2024, Vancouver,
                  BC, Canada, December 10 - 15, 2024},
  year         = {2024},
  url          = {http://papers.nips.cc/paper\_files/paper/2024/hash/ae0e43289bffea0c1fa34633fc608e92-Abstract-Datasets\_and\_Benchmarks\_Track.html},
  bibsource    = {dblp computer science bibliography, https://dblp.org},
  doi          = {10.52202/079017-3041},
  pages        = {95963--96010}
}

@misc{qwen3,
  title        = {Qwen3 Technical Report},
  author       = {An Yang and Anfeng Li and Baosong Yang and Beichen Zhang and Binyuan Hui and Bo Zheng and Bowen Yu and Chang Gao and Chengen Huang and Chenxu Lv and Chujie Zheng and Dayiheng Liu and Fan Zhou and Fei Huang and Feng Hu and Hao Ge and Haoran Wei and Huan Lin and Jialong Tang and Jian Yang and Jianhong Tu and Jianwei Zhang and Jianxin Yang and Jiaxi Yang and Jing Zhou and Jingren Zhou and Junyang Lin and Kai Dang and Keqin Bao and Kexin Yang and Le Yu and Lianghao Deng and Mei Li and Mingfeng Xue and Mingze Li and Pei Zhang and Peng Wang and Qin Zhu and Rui Men and Ruize Gao and Shixuan Liu and Shuang Luo and Tianhao Li and Tianyi Tang and Wenbiao Yin and Xingzhang Ren and Xinyu Wang and Xinyu Zhang and Xuancheng Ren and Yang Fan and Yang Su and Yichang Zhang and Yinger Zhang and Yu Wan and Yuqiong Liu and Zekun Wang and Zeyu Cui and Zhenru Zhang and Zhipeng Zhou and Zihan Qiu},
  year         = {2025},
  eprint       = {2505.09388},
  archivePrefix= {arXiv},
  primaryClass = {cs.CL},
  doi          = {10.48550/arXiv.2505.09388},
  url          = {https://arxiv.org/abs/2505.09388}
}

@misc{qwen25,
  title        = {Qwen2.5 Technical Report},
  author       = {{Qwen Team}},
  year         = {2025},
  eprint       = {2412.15115},
  archivePrefix= {arXiv},
  primaryClass = {cs.CL},
  doi          = {10.48550/arXiv.2412.15115},
  url          = {https://arxiv.org/abs/2412.15115}
}

@misc{qwen3vl,
  title        = {{Qwen3-VL} Technical Report},
  author       = {Shuai Bai and Yuxuan Cai and Ruizhe Chen and Keqin Chen and Xionghui Chen and Zesen Cheng and Lianghao Deng and Wei Ding and Chang Gao and Chunjiang Ge and Wenbin Ge and Zhifang Guo and Qidong Huang and Jie Huang and Fei Huang and Binyuan Hui and Shutong Jiang and Zhaohai Li and Mingsheng Li and Mei Li and Kaixin Li and Zicheng Lin and Junyang Lin and Xuejing Liu and Jiawei Liu and Chenglong Liu and Yang Liu and Dayiheng Liu and Shixuan Liu and Dunjie Lu and Ruilin Luo and Chenxu Lv and Rui Men and Lingchen Meng and Xuancheng Ren and Xingzhang Ren and Sibo Song and Yuchong Sun and Jun Tang and Jianhong Tu and Jianqiang Wan and Peng Wang and Pengfei Wang and Qiuyue Wang and Yuxuan Wang and Tianbao Xie and Yiheng Xu and Haiyang Xu and Jin Xu and Zhibo Yang and Mingkun Yang and Jianxin Yang and An Yang and Bowen Yu and Fei Zhang and Hang Zhang and Xi Zhang and Bo Zheng and Humen Zhong and Jingren Zhou and Fan Zhou and Jing Zhou and Yuanzhi Zhu and Ke Zhu},
  year         = {2025},
  eprint       = {2511.21631},
  archivePrefix= {arXiv},
  primaryClass = {cs.CV},
  doi          = {10.48550/arXiv.2511.21631},
  url          = {https://arxiv.org/abs/2511.21631}
}

@misc{internvl,
      title={{InternVL3.5}: Advancing Open-Source Multimodal Models in Versatility, Reasoning, and Efficiency}, 
      author={Weiyun Wang and Zhangwei Gao and Lixin Gu and Hengjun Pu and Long Cui and Xingguang Wei and Zhaoyang Liu and Linglin Jing and Shenglong Ye and Jie Shao and Zhaokai Wang and Zhe Chen and Hongjie Zhang and Ganlin Yang and Haomin Wang and Qi Wei and Jinhui Yin and Wenhao Li and Erfei Cui and Guanzhou Chen and Zichen Ding and Changyao Tian and Zhenyu Wu and Jingjing Xie and Zehao Li and Bowen Yang and Yuchen Duan and Xuehui Wang and Zhi Hou and Haoran Hao and Tianyi Zhang and Songze Li and Xiangyu Zhao and Haodong Duan and Nianchen Deng and Bin Fu and Yinan He and Yi Wang and Conghui He and Botian Shi and Junjun He and Yingtong Xiong and Han Lv and Lijun Wu and Wenqi Shao and Kaipeng Zhang and Huipeng Deng and Biqing Qi and Jiaye Ge and Qipeng Guo and Wenwei Zhang and Songyang Zhang and Maosong Cao and Junyao Lin and Kexian Tang and Jianfei Gao and Haian Huang and Yuzhe Gu and Chengqi Lyu and Huanze Tang and Rui Wang and Haijun Lv and Wanli Ouyang and Limin Wang and Min Dou and Xizhou Zhu and Tong Lu and Dahua Lin and Jifeng Dai and Weijie Su and Bowen Zhou and Kai Chen and Yu Qiao and Wenhai Wang and Gen Luo},
      year={2025},
      journal={CoRR},
      eprint={2508.18265},
      archivePrefix={arXiv},
      primaryClass={cs.CV},
      doi={10.48550/arXiv.2508.18265},
      url={https://arxiv.org/abs/2508.18265},
}

@inproceedings{osagents,
  title     = {{OS} Agents: A Survey on {MLLM}-based Agents for Computer, Phone and Browser Use},
  author    = {Hu, Xueyu and Xiong, Tao and Yi, Biao and Wei, Zishu and Xiao, Ruixuan and Chen, Yurun and Ye, Jiasheng and Tao, Meiling and Zhou, Xiangxin and Zhao, Ziyu and Li, Yuhuai and Xu, Shengze and Wang, Shenzhi and Xu, Xinchen and Qiao, Shuofei and Wang, Zhaokai and Kuang, Kun and Zeng, Tieyong and Wang, Liang and Li, Jiwei and Jiang, Yuchen Eleanor and Zhou, Wangchunshu and Wang, Guoyin and Yin, Keting and Zhao, Zhou and Yang, Hongxia and Wu, Fan and Zhang, Shengyu and Wu, Fei},
  booktitle = {Proceedings of the 63rd Annual Meeting of the Association for Computational Linguistics (Volume 1: Long Papers)},
  year      = {2025},
  address   = {Vienna, Austria},
  publisher = {Association for Computational Linguistics},
  url       = {https://aclanthology.org/2025.acl-long.369/},
  doi       = {10.18653/v1/2025.acl-long.369},
  pages     = {7436--7465}
}

@inproceedings{toolformer,
  author       = {Timo Schick and
                  Jane Dwivedi{-}Yu and
                  Roberto Dess{\`{\i}} and
                  Roberta Raileanu and
                  Maria Lomeli and
                  Eric Hambro and
                  Luke Zettlemoyer and
                  Nicola Cancedda and
                  Thomas Scialom},
  editor       = {Alice Oh and
                  Tristan Naumann and
                  Amir Globerson and
                  Kate Saenko and
                  Moritz Hardt and
                  Sergey Levine},
  title        = {Toolformer: Language Models Can Teach Themselves to Use Tools},
  booktitle    = {Advances in Neural Information Processing Systems 36: Annual Conference
                  on Neural Information Processing Systems 2023, NeurIPS 2023, New Orleans,
                  LA, USA, December 10 - 16, 2023},
  year         = {2023},
  url          = {http://papers.nips.cc/paper\_files/paper/2023/hash/d842425e4bf79ba039352da0f658a906-Abstract-Conference.html},
  bibsource    = {dblp computer science bibliography, https://dblp.org},
  doi          = {10.52202/075280-2997},
  pages        = {68539--68551}
}

@inproceedings{arigraph,
  author       = {Petr Anokhin and
                  Nikita Semenov and
                  Artyom Y. Sorokin and
                  Dmitry Evseev and
                  Andrey Kravchenko and
                  Mikhail Burtsev and
                  Evgeny Burnaev},
  title        = {{AriGraph}: Learning Knowledge Graph World Models with Episodic Memory
                  for {LLM} Agents},
  booktitle    = {Proceedings of the Thirty-Fourth International Joint Conference on
                  Artificial Intelligence, {IJCAI} 2025, Montreal, Canada, August 16-22,
                  2025},
  pages        = {12--20},
  publisher    = {ijcai.org},
  year         = {2025},
  url          = {https://doi.org/10.24963/ijcai.2025/2},
  doi          = {10.24963/IJCAI.2025/2},
  bibsource    = {dblp computer science bibliography, https://dblp.org}
}

@inproceedings{mango,
  title     = {{MANGO}: A Benchmark for Evaluating Mapping and Navigation Abilities of Large Language Models},
  author    = {Ding, Peng and Fang, Jiading and Li, Peng and Wang, Kangrui and Zhou, Xiaochen and Yu, Mo and Li, Jing and Walter, Matthew R. and Mei, Hongyuan},
  booktitle = {Proceedings of the First Conference on Language Modeling (COLM)},
  year      = {2024},
  url       = {https://openreview.net/forum?id=6vEfyp0o68}
}

@inproceedings{graphrectification,
  title     = {Constructing Coherent Spatial Memory in {LLM} Agents through Graph Rectification},
  author    = {Zhang, Puzhen and Chen, Xuyang and Feng, Yu and Jiang, Yuhan and Meng, Liqiu},
  booktitle = {Proceedings of the 64th Annual Meeting of the Association for Computational Linguistics (Volume 1: Long Papers)},
  year      = {2026},
  address   = {San Diego, California, USA},
  publisher = {Association for Computational Linguistics},
  url       = {https://aclanthology.org/2026.acl-long.2222/},
  doi       = {10.18653/v1/2026.acl-long.2222},
  pages     = {48126--48146}
}

@misc{gpt51,
  author = {{OpenAI}},
  title  = {Introducing {GPT-5.1} for Developers},
  year   = {2025},
  url    = {https://openai.com/index/gpt-5-1-for-developers/},
  note   = {Accessed: 2026-08-26}
}

@misc{gpt41,
  author = {{OpenAI}},
  title  = {Introducing {GPT-4.1} in the {API}},
  year   = {2025},
  url    = {https://openai.com/index/gpt-4-1/},
  note   = {Accessed: 2026-08-26}
}
